\documentclass[twocolumn]{svjour3}

\usepackage{graphicx}%
\usepackage{multirow}%
\usepackage{amsmath,amssymb,amsfonts}%

\usepackage{amsthm}%
\usepackage{mathrsfs}%
\usepackage[title]{appendix}%
\usepackage{xcolor}%
\usepackage{textcomp}%
\usepackage{manyfoot}%
\usepackage{booktabs}%
\usepackage{algorithm}
\usepackage{algorithmic}%
\usepackage{listings}%
\usepackage{bm}
\usepackage{wrapfig}
\usepackage{authblk}
\usepackage[misc]{ifsym}

\begin{document}
\onecolumn
\pagenumbering{gobble}
\Large
\begin{center}
PRCL: Probabilistic Representation Contrastive Learning for Semi-Supervised Semantic Segmentation\\ 

\hspace{10pt}

\large
Haoyu Xie$^{1}$, Changqi Wang$^2$, Jian Zhao$^3$, Yang Liu$^4$, Jun Dan$^5$, Chong Fu$^6$, Baigui Sun$^7$\\

\hspace{10pt}

\small  
$1$) First author, 2010643@stu.neu.edu.cn, School of Computer Science and Engineering, Northeastern University, Shenyang, 110819, China\\
$2$) First author, 2101668@stu.neu.edu.cn, School of Computer Science and Engineering, Northeastern University, Shenyang, 110819, China\\
$3$) First author, Corresponding author, zhaojian90@u.nus.edu, Intelligent Game and Decision Laboratory, Beijing 100191, China \\
$4$) ly261666@alibaba-inc.com, Alibaba DAMO Academy, Alibaba Group, Hangzhou 310000, China\\
$5$) danjun@zju.edu.cn, Department of Information Science and Electronic Engineering, Zhejiang University, Hangzhou 310027, China\\
$6$) fuchong@mail.neu.edu.cn, School of Computer Science and Engineering, Northeastern University, Shenyang, 110819, China\\
$7$) Corresponding author, baigui.sbg@alibaba-inc.com , Alibaba DAMO Academy, Alibaba Group, Hangzhou 310000, China\\

\end{center}

\hspace{10pt}

\normalsize

\paragraph{Abstract}
Tremendous breakthroughs have been developed in Semi-Supervised Semantic Segmentation (S4) through contrastive learning.
However, due to limited annotations, the guidance on unlabeled images is generated by the model itself, which inevitably exists noise and disturbs the unsupervised training process.
To address this issue, we propose a robust contrastive-based S4 framework, termed the Probabilistic Representation Contrastive Learning (PRCL) framework to enhance the robustness of the unsupervised training process.
We model the pixel-wise representation as Probabilistic Representations (PR) via multivariate Gaussian distribution and tune the contribution of the ambiguous representations to tolerate the risk of inaccurate guidance in contrastive learning.
Furthermore, we introduce Global Distribution Prototypes (GDP) by gathering all PRs throughout the whole training process.
Since the GDP contains the information of all representations with the same class, it is robust from the instant noise in representations and bears the intra-class variance of representations.
In addition, we generate Virtual Negatives (VNs) based on GDP to involve the contrastive learning process.
Extensive experiments on two public benchmarks demonstrate the superiority of our PRCL framework.
		
\paragraph{\small{Key Word}}
		Semi-Supervised Semantic Segmentation; 
		Contrastive Learning;
		Probabilistic Representation; 
		Robust Learning; 
  
\paragraph{Statements and Declarations}
The authors declare that they have no known competing financial interests or personal relationships that could have appeared to influence the work reported in this paper.

\clearpage
\twocolumn

\title{PRCL: Probabilistic Representation Contrastive Learning for Semi-Supervised Semantic Segmentation}


\author{Haoyu Xie$^{1, 3, \ast}$ \thanks{$^{\ast}$ equal contribution}\and
Changqi Wang$^{1, \ast}$ \and
Jian Zhao$^{2, \ast}$ \Letter \thanks{\Letter Corresponding author} \and
Yang Liu$^3$ \and
Jun Dan$^4$ \and
Chong Fu$^1$ \and 
Baigui Sun$^{3}$ \Letter
}

\institute{
1. School of Computer Science and Engineering, Northeastern University, Shenyang 110819, China \\ 
2. Intelligent Game and Decision Laboratory, Beijing 100191, China  \\ 
3. Alibaba DAMO Academy, Alibaba Group, Hangzhou 310000, China \\ 
4. Department of Information Science and Electronic Engineering, Zhejiang University, Hangzhou
          310027, China
          }

\date{Received: date / Accepted: date}

\maketitle
\abstract{
Tremendous breakthroughs have been developed in Semi-Supervised Semantic Segmentation (S4) through contrastive learning.
However, due to limited annotations, the guidance on unlabeled images is generated by the model itself, which inevitably exists noise and disturbs the unsupervised training process.
To address this issue, we propose a robust contrastive-based S4 framework, termed the Probabilistic Representation Contrastive Learning (PRCL) framework to enhance the robustness of the unsupervised training process.
We model the pixel-wise representation as Probabilistic Representations (PR) via multivariate Gaussian distribution and tune the contribution of the ambiguous representations to tolerate the risk of inaccurate guidance in contrastive learning.
Furthermore, we introduce Global Distribution Prototypes (GDP) by gathering all PRs throughout the whole training process.
Since the GDP contains the information of all representations with the same class, it is robust from the instant noise in representations and bears the intra-class variance of representations.
In addition, we generate Virtual Negatives (VNs) based on GDP to involve the contrastive learning process.
Extensive experiments on two public benchmarks demonstrate the superiority of our PRCL framework.
}

\keywords{Semi-Supervised Semantic Segmentation; Contrastive Learning; Probabilistic Representation;  Robust Learning}



\maketitle

\section{Introduction}\label{intro}
Semantic Segmentation is a fundamental task in computer vision, aiming to predict the class of each pixel.
Significant progress has been made via training a segmentation model \cite{FCN,deeplabv3+,segformer} on large-scale annotated images, which requires high labor costs.
Semi-Supervised Semantic Segmentation (S4) leverages unlabeled images in the training process to further improve the performance of the segmentation model via adversarial training \cite{adversarial4s4}, consistency regularization \cite{cotraining}, and self-training \cite{MT}, which ease the thirsty of annotated images.
\begin{figure*}[t]
  \centering
  \includegraphics[width=0.75\linewidth]{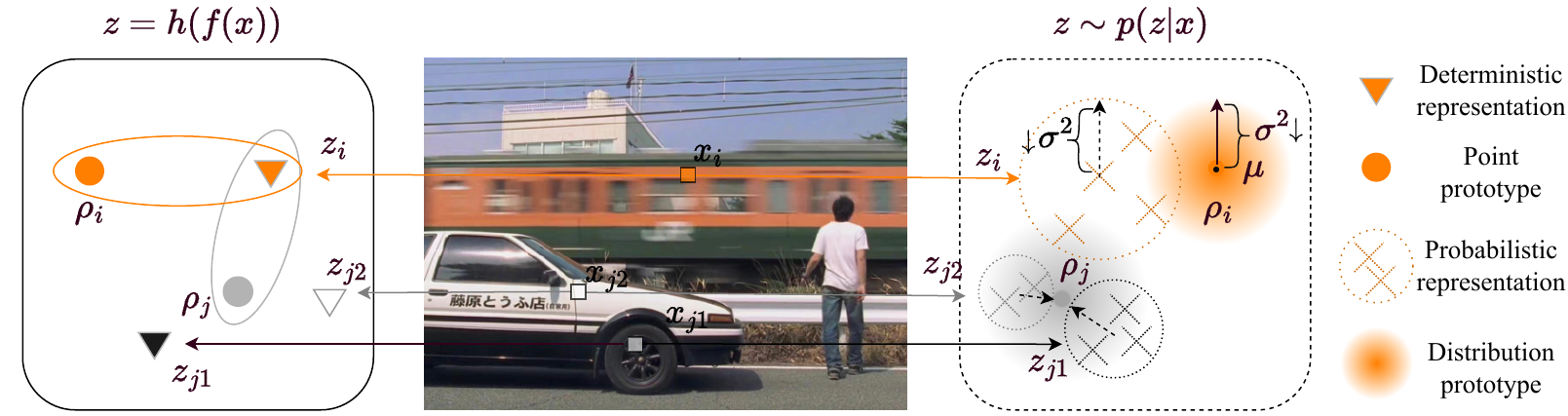}
  \caption{Contradistinction between two types of representations and prototypes. Point prototype means the prototype of the deterministic representation and distribution prototype means the prototype of the probabilistic representation. Distinct from conventional representation, we introduce probability, and thus regarding representation as a multivariate Gaussian distribution. The probabilistic representation is able to demolish the ambiguity of representation prototype mapping to some extent and enhance the robustness of the model during training fuzzy pixels.
   }
  \label{probabilistic_rep}
\end{figure*}

Self-training is a well-known paradigm extensively employed in S4 tasks.
It involves leveraging a model pre-trained on labeled images to generate predictions, \textit{a.k.a}, pseudo-labels, for unlabeled images.
These pseudo-labels in conjunction with annotations are subsequently used as guidance to retrain the model.
Recent powerful methods introduce the concept of pixel-wise contrastive learning to the self-training paradigm, aiming to explore the semantic information not only in the local context of a single image but also in the images in a mini-batch or even the entire dataset.
This is achieved by projecting pixels to representations in the latent space, where representations from the same class are aggregated around their class centroid, \textit{a.k.a}, prototype, while those from different classes, \textit{a.k.a}, negative representations, are separated.
In the context of semi-supervised learning, the semantic guidance of unlabeled images for contrastive learning comes from the pseudo-labels during training.
Consequently, the quality of pseudo-labels is critical in contrastive learning since inaccurate pseudo-labels lead to assigning representations to wrong classes and cause a disorder in latent space.
Existing methods have attempted to refine pseudo-labels via their corresponding confidence \cite{Reco} or entropy \cite{DMT}.
While these techniques have shown promise in enhancing the quality of pseudo-labels and eliminating inaccurate ones to some extent, they rely on delicate strategies and still struggle to fully address the inherent noise and incorrectness in pseudo-labels.
Motivated by these challenges, our goal is to improve the robustness of representations, enabling them to perform more effectively even in the presence of inaccurate pseudo-labels.
By focusing on enhancing the robustness of representations to accommodate imperfect guidance, we are able to enhance the overall performance and reliability of contrastive-based S4 approaches.

In contrast to existing conventional \textit{deterministic} representation modeling, which maps the representation to the deterministic point in the latent space, our proposed method introduces a novel perspective by treating representations as random variables with learnable parameters, termed Probabilistic Representation (PR).
Specifically, we adopt a multivariate Gaussian distribution to model the representations, thereby obtaining distribution prototypes.
As illustrated in Fig.~\ref{probabilistic_rep}, this probabilistic modeling is reflected in the expression $z \sim p(z|x)$.
The pixel of the fuzzy train carriage $x_i$ is mapped to $z_i$ in the latent space which encompasses two components: the most likely representation $\mu$ (mean) and the probability $\sigma^2$ (variance) of the distribution.
Similarly, the pixels of the car $x_{j1}$ and $x_{j2}$ are mapped to $z_{j1}$ and $z_{j2}$ respectively.
For comparison, deterministic mapping is shown in $z=h(f(x))$.
In scenarios where the distance between the representation $z_i$ to prototype $\rho_i$ is the same as the distance from $z_i$ to $\rho_j$, deterministic representation encounters an ambiguity in mapping $z_i$ to either $\rho_i$ or $\rho_j$.
On the contrary, in the latent space of probabilistic representations, $z_i$ is mapped to $\rho_i$ since $\rho_i$ possesses a smaller value of $\sigma^2$ compared to $\rho_j$.
It is worth noting that $\sigma^2$ is inversely proportional to the probability, indicating that the mapping from $z_i$ to $\rho_i$ is more reliable than mapping to $\rho_j$ according to the probability.

\begin{figure*}[t]
  \centering
  \includegraphics[width=0.8\linewidth]{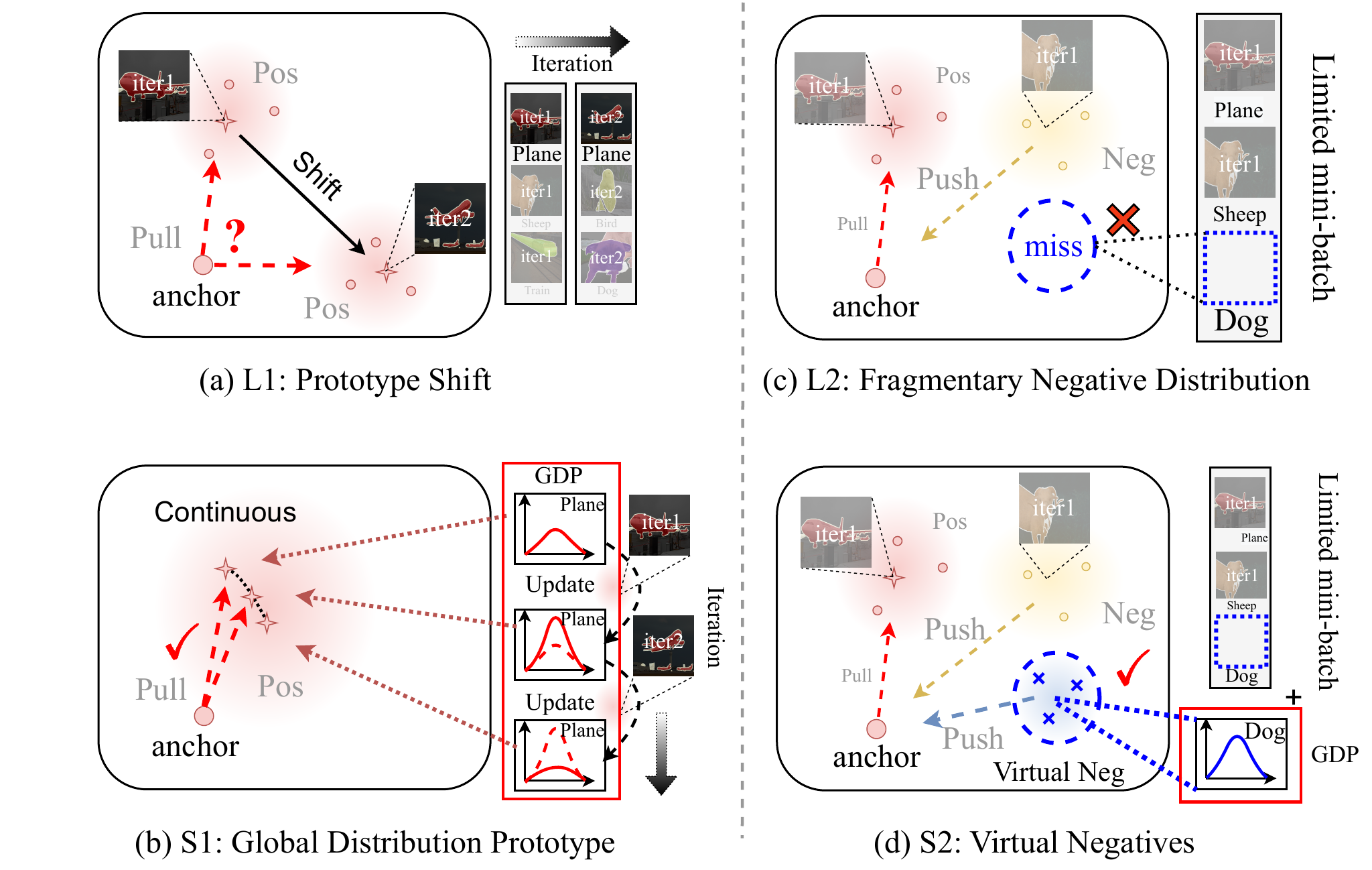}
  \caption{Our PRCL framework tackles the negative effect brought by prototype shift and fragmentary negative distribution (L1 and L2) with our proposed global distribution prototype and virtual negatives. (S1 and S2).
   }
  \label{coverfig}
\end{figure*}
Meanwhile, recent contrastive-based S4 works suffer from the limitation of solely considering the contrast in the current iteration.
Specifically, the prototypes in those methods are obtained by aggregating the semantic information of representations belonging to the same class in the current iteration.
This approach can result in prototype shifts across adjacent iterations due to discrepancies among representations caused by inaccurate pseudo-labels and intra-class variance.
We argue that prototype consistency is crucial to establish a stable direction for representation aggregation.
In addition, since negative representations are from the current mini-batch when solely considering the contrast in the current iteration, the distribution of negative representations is fragmentary due to the limited size of the mini-batch.
Some approaches \cite{CLCMB,RegionContrast} leverage the memory bank strategy to compensate for the fragmentary distribution, which stores representations in the external memory and sample from them when constructing negative distribution.
However, dense pixel-wise representations lead to significant memory overhead and high computational cost.
To overcome the above limitations, we rethink pixel-wise contrastive learning from the global perspective and build Global Distribution Prototypes (GDP) based on probabilistic representations.
Distinct from the conventional prototypes which represent the semantic information of identical class representations in the current mini-batch, GDPs aggregate the representations along training iterations and are updated with the prototypes of the current iteration as an observation of Bayesian Estimation \cite{Bayesian}.
By employing GDPs to bridge iterations during the training process, our method enables prototypes to withstand instant noise in representations and accommodate intra-class variance among identical class representations.
Therefore, the prototype is more consistent across iterations and provides a stable direction for identical representation aggregation.
In contrast to the conventional approach of constructing a memory bank to provide a large number of negatives, we propose a novel strategy called Virtual Negatives (VN).
By leveraging a reparameter trick from the GDPs, we generate VNs and facilitate a balance between compactness and diversity of them through a virtual radius.
Notably, compared with the typical memory bank solution, our VN reduces the GPU memory from 2.63GB to 42KB, and also accelerates the training process for 22.23\%.
Most importantly, our introduced VN strategy performs better than the conventional memory bank strategy.

This work is based on our previous conference version [1] by tackling the limitations of merely considering the contrast in the current iteration, including the prototype shift between adjacent iterations and fragmentary negative distribution. Specifically, we propose GDP with an update strategy and VN to maintain the prototype consistency and compensate for the fragmentary negative distribution, respectively.
To summarize, our main contributions are four-fold:
\begin{itemize}
    \item We introduce the \textit{probabilistic representation} and improve the robustness of representations in contrastive learning, which eases the negative effects of inaccurate pseudo-labels.
    \item We build \textit{global distributional prototypes} and \textit{virtual negatives} to make up for the defects brought by the limited size of the mini-batch, which is more memory efficient and faster than the conventional memory bank strategy.
    \item Extensive experiments on PASCAL VOC 2012 and Cityscapes demonstrate the effectiveness of our proposed method.
    \item We present comprehensive ablation studies and in-depth analysis of probabilistic representation, global distributional, and virtual negatives, which demonstrate that our method not only improves the robustness and performance of segmentation model in S4.
\end{itemize}

\section{Related Work}\label{related work}
\subsection{Semi-supervised Semantic Segmentation}
Semantic segmentation aims to classify each pixel in an entire image by class.
Training models for this task typically requires a substantial amount of labeled data, involving meticulous manual annotations. Semi-supervised learning methods have emerged as effective approaches to leverage large volumes of unlabeled data, thereby reducing the dependency on extensive manual annotations.
Self-training \cite{adversarial4s4,3stage_selftraining,ST++,fuzzy_positive} and consistency regularization \cite{CCT,cotraining,TCC,UCC} are two widely-used paradigms.
Self-training methods leverage high-dimensional perturbations \cite{cutmix,classmix,AEL} and refined pseudo-labels \cite{FixMatch,DMT} to enhance their performance.
Some methods \cite{adversarial4s4,universal_s4,ss_with_gm,s4_with_consistency} based on GANs \cite{gan} and adversarial learning \cite{vat} concentrate on generating more ground-truth like predictions.
Additionally, methods focusing on balancing class distribution \cite{subclass_regular,DARS,BBN,CReST} have demonstrated competitiveness in specific scenarios. 
Recent works based on self-training \cite{Reco,U2PL,SePiCo} emphasize the regularization of representations in the latent space to maintain an ordered latent space.
This improves the quality of features and ultimately boosts model performance, which is also our goal.
\subsection{Pixel-wise Contrastive Learning}
Distinct from instance-wise contrastive learning \cite{InstDisc,yemang_InstFeat,SimCLR,MoCo,BYOL,cite1}, which treats each image as an individual class and distinguishes it from other images through multiple views.
In the case of pixel-wise contrastive learning \cite{s4_with_context,RegionContrast,Cipc,PixPro,cite2,space_engage,cite3}, dense pixel-wise representations are distinguished by semantic guidance, \textit{i.e.}, labels or pseudo-labels.
However, in the semi-supervised setting, the availability of labeled images is limited, and the majority of pixel classifications rely on pseudo-labels, which can introduce inaccuracies and disrupt the latent space.
To address these challenges, previous methods \cite{Reco,CLCMB,U2PL} have attempted to refine pseudo-labels using threshold sampling strategies.
In contrast, our approach focuses on improving the quality of representations and accommodating inaccurate pseudo-labels, rather than solely filtering them out.
By emphasizing the enhancement of representation quality, we aim to ease the negative effects of inaccurate pseudo-labels and foster a more robust and ordered latent space.

The process of pixel-wise contrastive learning is to aggregate representations of the same class to their prototype (class centroid) and separate them away from negative representations (representations with different classes).
Due to limited GPU memory, the prototype merely gathers the semantic information of representation in the current iteration in most methods \cite{Reco,U2PL}, thereby disregarding the global semantic information of the entire dataset.
To address this limitation, some methods have proposed the use of a memory bank to store representations from past iterations \cite{RegionContrast} or update the prototype using Exponential Moving Average (EMA) \cite{proto_consistency,Rethinking_semantic_segmentation}.
In our approach, we introduce a novel strategy that considers all historical representations, overcoming the limitation.
As for negative representations, some methods \cite{Reco} sample them from the current iteration.
However, due to the limited batch size, representations in the current iteration may not cover all classes, leading to a fragmentary negative distribution problem.
To mitigate this issue, some methods try to alleviate it by introducing the memory bank \cite{density_guided,CLCMB,U2PL} or approximating the ideal negative distribution through a probabilistic way \cite{SePiCo}.
In our method, we compensate for the negative distribution by generating representations, which takes little memory consumption and minimal computational cost.
\subsection{Probabilistic Embedding}
Probabilistic Embedding (PE) extends the concept of conventional embeddings by predicting the overall distribution of embeddings \textit{e.g.}, Gaussian \cite{PFE} and von Mises-Fisher \cite{SCF}, instead of a single vector.
The ability of neural networks to predict distributions stems from the work of Mixture Density Networks (MDN) \cite{MDN}.
Variable Auto-Encoders (VAE) \cite{VAE} introduced the use of MLP to predict the mean and variance of a distribution, which serves as the foundation for many probabilistic embedding approaches \cite{PFE,Stochastic_prototype_embeddings,vMF_loss,Probabilistic_representations_for_video_contrastive_learning}.
Hedged Instance emBeddings (HIB) \cite{HIB} attempt to apply PE to image retrieval and verification tasks.
Subsequently, PE is applied to face verification tasks.
Probabilistic Face Embeddings (PFE) \cite{PFE} maps each image to a Gaussian distribution in the latent space, with the mean predicted by a pre-trained model and the variance estimated through an MLP (probability head). 
Sphere Confidence Face (SCF) \cite{SCF} maps images to a von Mises-Fisher distribution with mean and concentration parameters.
We adopt a similar architecture, but it is important to note that PFE and SCF optimize the mean and variance (concentration parameter) in two separate stages.
Specifically, they pre-train a deterministic model to predict the mean and then freeze it while optimizing the variance.
In our work, however, we optimize the mean and variance simultaneously, enabling them to interact with each other.

While conventional distance metrics measure the similarity between distributions based on their means, they are inadequate for capturing probabilistic similarity due to the variance component.
To address this challenge, HIB employs the reparameter trick \cite{VAE} to obtain two sets of samples from two distributions through Monte-Carlo sampling, and accumulates the similarity of samples to represent the similarity of distributions.
In contrast, PFE and SCF directly compute distribution similarity using the mutual likelihood score.
However, these methods are unable to optimize the mean and variance simultaneously since uncertainty/probability (variance) is only informative if representation (mean) is reasonable.
PFE and SCF tackle this issue by training the mean and variance in two separate stages.
In our approach, we address this by training the mean and variance separately with different learning rates.
\section{Methodology}
In the S4 task, we can achieve a small labeled set $\mathcal{D}_l=\{(\bm{x}^l_i,\bm{y}^l_i)\}_{i=1}^{N_l}$ and a large unlabeled set $\mathcal{D}_u=\{\bm{x}^u_i\}_{i=1}^{N_u}$, where labeled dataset $\mathcal{D}_l$ contains $N_l$ pairs of images and corresponding pixel-wise labels $(\bm{x}^l, \bm{y}^l)$ and unlabeled dataset only contains $N_u$ unlabeled images.
Our goal is to train a segmentation model with $\mathcal{D}_l$ and $\mathcal{D}_u$.
The base segmentation model contains an encoder $f(\cdot)$ and a segmentation head $g(\cdot)$.
We adopt the teacher-student paradigm and pixel-wise contrastive learning to our framework, described in Sec.~\ref{mt}.

\subsection{MT and Contrastive Learning}\label{mt}
The standard teacher-student paradigm consists of two segmentation models with the same architecture, named the student model and the teacher model respectively.
We denote $f(\cdot)$, $g(\cdot)$ as the encoder and segmentation head in the student model and $f'(\cdot)$, $g'(\cdot)$ as those of the teacher model.
The student model parameters are optimized via Stochastic Gradient Descent (SGD) to minimize the loss function $\mathcal{L}$ while the parameters in the teacher model are updated by the Exponential Moving Average (EMA) of the parameters in the student model.
The pseudo-labels $\bm{y}^u_i$ for training the student model during the unsupervised process are produced based on the output logits from the teacher model, \textit{i.e.}, $\bm{p}^u_i = g'(f'(\bm{x}^u_i))$, formulated as:
\begin{equation}\label{eq1}
    \scalebox{0.9}{$ \displaystyle
    \begin{aligned}
        \bm{y}_i^{u} = \bm{1}_c(\mathop{\arg\max}\limits_{c}\{p_{i,c}^u\}_{c\in C}),
    \end{aligned}
    $}
\end{equation}
where $p_{i,c}^u$ denotes value of $\bm{p}^u_i$ on $c^{th}$ dimension,  $\bm{1}_c(\cdot)$ denotes the one-hot encoding of class $c$ and $C$ denotes the set of total classes in the dataset.
In order to additionally regularize the model's output in the latent space, recent methods \cite{CLCMB,Reco,U2PL,Cipc,RegionContrast,proto_contrast_da_s2,density_guided} introduce contrastive learning to the teacher-student paradigm.
Concretely, a representation head $h(\cdot)$ is introduced to the student model to map pixels to representations, \textit{i.e.}, $\bm{z_i}=h(f(\bm{x}_i))$.
The representations of the identical class are aggregated to their prototype $\bm{\rho}$ and those of different classes are separated via a contrastive loss (\textit{e.g.}, InfoNCE).
The prototype $\bm{\rho}$ of each class is obtained by gathering the semantic information of the identical class representations.
The semantic guidance of contrastive learning is also from $\bm{y}^l_i$ and $\bm{y}^u_i$.

\noindent\textbf{Discussion.}
In this paper, we argue that recent S4 models based on pixel-wise contrastive learning suffer from two potential limitations:
\textbf{1)} \textit{The model suffers from poor robustness in the case of inaccurate pseudo-labels}.
Since the pseudo-labels $\bm{y}^u_i$ are derived solely from the prediction of the teacher model's segmentation head $g'(\cdot)$, there exist inaccurate ones due to the limited cognitive ability of the teacher model.
Even though the quality of pseudo-labels can be improved by adopting delicate sampling strategies, the essential errors in pseudo-labels are rather hard to be eliminated.
Those inaccurate pseudo-labels will mislead the model if they are directly applied as the supervision during training on $\mathcal{D}_u$.
The strategy of learning a robust contrastive-based S4 model under the inaccurate pseudo-labels has been overlooked and remains unexplored.
\textbf{2)} \textit{The prototype is shifted and the negative distribution is fragmentary}.
In recent works, the prototype $\bm{\rho}$ is calculated as the mean of the identical class representations in the current iteration.
However, due to incorrect pseudo-labels and intra-class variance, the representations of the same class can significantly vary across different iterations, resulting in the position of the prototype changing dramatically, termed prototype shift.
We argue that this shift in the prototype hinders the provision of a consistent direction for aggregating identical representations and leads to a disordered latent space.
Additionally, since the negative representations are from the current mini-batch, the limited mini-batch size leads to a fragmentary distribution of negatives for the contrastive learning process within the current iteration.
While some approaches attempt to address this limitation by employing a memory bank to store representations from past iterations and sample negatives from it, this strategy incurs high memory usage and computational costs.

To mitigate these limitations, we build a framework that is robust against inaccurate pseudo-labels.
The representations in our framework are modeled via Gaussian distribution, described in Sec.~\ref{pr}.
We build Global Distribution Prototypes (GDP) to maintain the consistency of prototypes in Sec.~\ref{gdp} and obtain Virtual Negatives (VN) based on GDPs for compensating fragmentary negative distribution, described in Sec.~\ref{vn_sec}.

\begin{figure*}[t]
  \centering
  \includegraphics[width=1.0\linewidth]{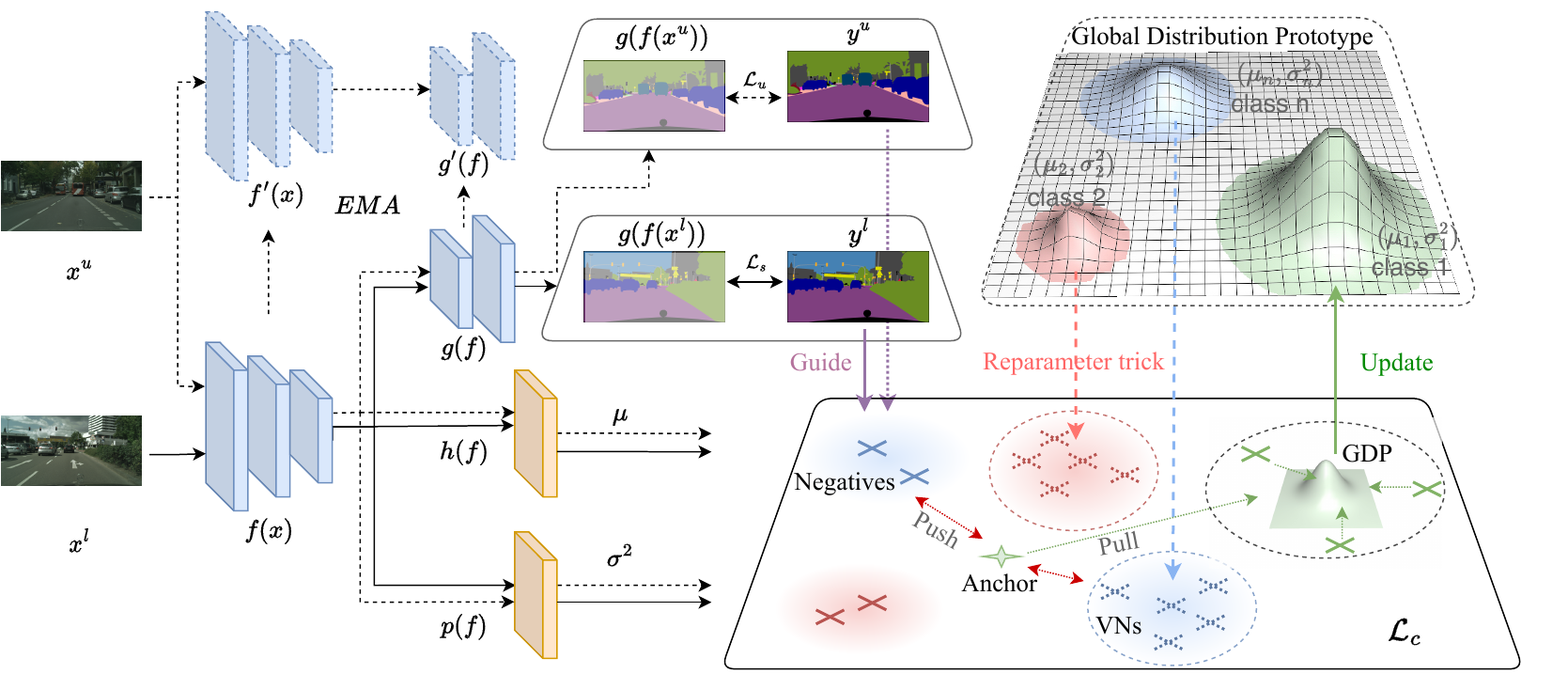}
  \caption{Overall framework of PRCL. The training pipeline contains two input streams: labeled images (black arrows) and unlabeled images (black dash arrows).
  In the pixel space, the model is guided by the combination of ground-truth $y^l$ and original pseudo-labels $y^u$.
  In the latent space, the model maps the pixels into probabilistic representations $\bm{z}\sim \mathcal{N}(\bm{\mu}, \bm{\sigma}^2)$ via two heads: $h(\cdot)$ and $p(\cdot)$.
  And the GDP is stored in a prototype-level dictionary and is updated with the local prototype.
  We generate the virtual negatives (VN, dashed cross) from GDP for contrastive loss $\mathcal{L}_{c}$.
   }
  \label{framework}
\end{figure*}

\subsection{Probabilistic Representation}\label{pr}
In this section, we detail the process of building our probabilistic representations and the similarity measurement for probabilistic representations.

We denote the probability of mapping a pixel $\bm{x}_i$ to a representation $\bm{z}_i$ as $p(\bm{z}_i|\bm{x}_i)$ and define the representation as a random variable following it.
For simplicity, we take the form of multivariate Gaussian distribution $\mathcal{N}(\bm{\mu}_i, \bm{\sigma}_i^2\bm{I})$ as:
\begin{equation}\label{eq2}
\scalebox{0.9}{$ \displaystyle
    \begin{aligned}
     p(\bm{z}_i|\bm{x}_i) = \mathcal{N}(\bm{z}; \bm{\mu}_i,\bm{\sigma}^2_i\bm{I}),
    \end{aligned}
    $}
\end{equation}
where $\bm{I}$ represents the unit diagonal matrix.
In this formulation, $\bm{\mu}$ represents the most likely values for the representation, while $\bm{\sigma}^2$ captures the associated probability.
It is worth noting that $\bm{\sigma}^2$ is inversely related to the probability, meaning that larger $\bm{\sigma}^2$ indicate lower probabilities.
Both $\bm{\mu}$ and $\bm{\sigma}^2$ have the same dimensions.
The mean $\bm{\mu}$ is predicted by the representation head $h(\cdot)$. Meanwhile, we introduce a probability head $p(\cdot)$ in parallel to predict the probability $\bm{\sigma}^2$.

In conventional contrastive learning, the similarity between representations is typically measured using the $\ell_2$ distance or cosine similarity, which does not possess the ability to quantify the similarity between two distributions.
To solve this problem, we employ the Mutual likelihood Score (MLS) as the measurement of the similarity between two distributions $\bm{z}_i$ and $\bm{z}_j$, as follows:
\begin{equation}\label{eq3}
\scalebox{0.9}{$ \displaystyle
    \begin{aligned}
        MLS(\bm{z}_i,\bm{z}_j)=& \log(p(\bm{z}_i=\bm{z}_j))\\
        =&-\frac{1}{2}\sum_{l=1}^D(\frac{(\mu_i^{(l)}-\mu_j^{(l)})^2}{\sigma_i^{2(l)}+\sigma_j^{2(l)}}\\
        &+ \log(\sigma_i^{2(l)}+\sigma_j^{2(l)}))-\frac{D}{2}log2\pi,
    \end{aligned}
    $}
\end{equation}
where $\mu_i^{(l)}$ denotes to the $l^{th}$ dimension of $\bm{\mu}_i$ and the same for $\sigma_i^{(l)}$.
MLS combines a weighted $\ell_2$ distance and a log regularization term, essentially.
The conventional $\ell_2$ distance solely considers the similarity between representations mapped in the latent space based on pseudo-labels, without taking into account their reliability.
However, inaccurate pseudo-labels can lead to incorrect optimization directions and disrupt the latent space.
To address this issue, the MLS incorporates the probabilities of $\bm{z}_i$ and $\bm{z}_j$ to account for inaccurate pseudo-labels from two perspectives:
\textbf{(\romannumeral1)}: In the first term, the weight of $\ell_2$ distance is reduced when the $\bm{\sigma}^2$ is large.
This indicates that the similarity between $\bm{z}_i$ and $\bm{z}_j$ decreases due to the low probabilities, even if their $\ell_2$ distance suggests they are similar.
By considering the probabilities, the MLS incorporates a measure of similarity that accounts for the reliability of representations.
\textbf{(\romannumeral2)}: In the second term, the log regularization penalizes low probability representations, which encourages all representations to be more reliable. 
Additionally, $\bm{\sigma}^2$ and $\bm{\mu}$ can interact with each other.
The learnable $\bm{\sigma}^2$ is associated with $\ell_2$ distance, allowing it to be learned based on the relationships among representations.
Conversely, the $\bm{\mu}$ can also be optimized via the $\bm{\sigma}^2$.
This mutual interaction between $\bm{\mu}$ and $\bm{\sigma}^2$ aligns with our intuitive understanding of representation learning.

\subsection{Global Distribution Prototype}\label{gdp}
In conventional methods, the prototype $\bm{\rho}_c$ of class $c$ is typically obtained by aggregating the semantic information from all representations that belong to class $c$ in the current iteration.
With probabilistic representations, this process can be formulated as:
\begin{equation}\label{eq4}
\scalebox{0.9}{$ \displaystyle
    \begin{aligned}
        &\bm{\rho}_c \sim \mathcal{N}(\hat{\bm{\mu}_c}, \hat{\bm{\sigma}}^2_c\bm{I}),\\
        & \frac{1}{\hat{\bm{\sigma}}^2_c} = \sum_{\bm{z}_{ci}\in \mathcal{\bm{Z}}_c}\frac{1}{\bm{\sigma}_{ci}^2},\\
        & \hat{\bm{\mu}_c} = \sum_{\bm{z}_{ci}\in \mathcal{\bm{Z}}_c} \frac{\hat{\bm{\sigma}}^2_c}{\bm{\sigma}_{ci}^2}\bm{\mu}_{ci},\\
    \end{aligned}
    $}
\end{equation}
where $\mathcal{Z}_c$ represents the set of the representations belong to class $c$ in current iteration $\mathcal{Z}_c=\{\bm{z}_{c0}, \bm{z}_{c1},...,\bm{z}_{ci}\}$ and $\bm{z}_{ci}\stackrel{i.i.d}{\sim}\mathcal{N}(\bm{\mu}_{ci}, \bm{\sigma}^2_{ci}\bm{I})$.
Even though probabilistic representations offer the advantage of accommodating inaccurate pseudo-labels by incorporating probability into the representation, certain significant inaccuracies still affect the precision of prototypes, resulting in prototype shift.
Additionally, the inherent intra-class variance introduces variations in the representations of identical classes across adjacent iterations, further causing the prototype shift. To address these challenges, we introduce an efficient strategy that sequentially aggregates representations across iterations from a global perspective.
Specifically, we define the prototype calculated in the current iteration as \textit{local} prototype and extend the local prototype to the Global Distribution Prototype (GDP).
We introduce a variable $t$ to represent the $t^{th}$ iteration during training and use $\bm{\rho}_l(t)$ to represent the \textit{local} prototype.
For clarity, we omit class $c$ here.
We represent GDP as $\bm{\rho}_g(t)$, which can be formulated as:
\begin{equation}\label{eq5}
\scalebox{0.9}{$ \displaystyle
    p(\bm{\rho}_g(t)|\mathcal{Z}_g(t))= \mathcal{N}(\hat{\bm{\mu}}_g(t), \hat{\bm{\sigma}}^2_g(t)\bm{I}),
    $}
\end{equation}
where $\mathcal{Z}_g(t)$ represents the set of all identical class representations observed, and given $\mathcal{Z}_l(t)$ represents the set of identical class representations in the $t^{th}$ iteration, $\mathcal{Z}_g(t)=\mathcal{Z}_l(0)\cup \mathcal{Z}_l(1)\ldots \cup \mathcal{Z}_l(t)$.
Since each representation $\bm{z}_i\stackrel{i.i.d}{\sim} \mathcal{N}(\bm{\mu}_i, \bm{\sigma}^2_i\bm{I})$, we have
\begin{equation}\label{eq6}
\scalebox{0.8}{$ \displaystyle
    \begin{aligned}
        \frac{1}{\hat{\bm{\sigma}}^2_g(t)} &= \sum_{\bm{z}_i\in \mathcal{\bm{Z}}_g(t)}\frac{1}{\bm{\sigma}_i^2}
        =\sum_{\bm{z}_i\in \mathcal{Z}_g(t-1)}\frac{1}{\bm{\sigma}^2_i}+\sum_{\bm{z}_i\in \mathcal{Z}_l(t)}\frac{1}{\bm{\sigma}^2_i}\\
        &=\frac{1}{\hat{\bm{\sigma}}^2_g(t-1)}+\frac{1}{\hat{\bm{\sigma}}^2_l(t)},
    \end{aligned}
    $}
\end{equation}
and
\begin{equation}\label{eq7}
    \scalebox{0.7}{$ \displaystyle
    \begin{aligned}
        \hat{\bm{\mu}}_g(t) &= \sum_{\bm{z}_i\in \mathcal{\bm{Z}}_g(t)} \frac{\hat{\bm{\sigma}}^2_g(t)}{\bm{\sigma}_i^2}\bm{\mu}_i\\
        &=\hat{\bm{\sigma}}^2_g(t)\Biggl(\sum_{\bm{z}_i\in \mathcal{Z}_g(t-1)}\frac{\bm{\mu}_i}{\bm{\sigma}_i^2}+\sum_{\bm{z}_i\in \mathcal{Z}_l(t)}\frac{\bm{\mu}_i}{\bm{\sigma}_i^2}\Biggl)\\
        &=\hat{\bm{\sigma}}^2_g(t)\Biggl(\frac{1}{\hat{\bm{\sigma}}^2_g(t-1)}\sum_{\bm{z}_i\in \mathcal{Z}_g(t-1)}\frac{\hat{\bm{\sigma}}^2_g(t-1)\bm{\mu}_i}{\bm{\sigma}_i^2}+\frac{1}{\hat{\bm{\sigma}}^2_l(t)}\sum_{\bm{z}_i\in \mathcal{Z}_l(t)}\frac{\hat{\bm{\sigma}}^2_l(t)\bm{\mu}_i}{\bm{\sigma}_i^2}\Biggl)\\
        &=\hat{\bm{\sigma}}^2_g(t)\Biggl(\frac{\hat{\bm{\mu}}_g(t-1)}{\hat{\bm{\sigma}}_g^2(t-1)}+\frac{\hat{\bm{\mu}}_l(t)}{\hat{\bm{\sigma}}^2_l(t)}\Biggl).
    \end{aligned}
    $}
\end{equation}
This means that we can get the current GDP from the last GDP $\bm{\rho}_g(t-1)$ and the current local prototype $\bm{\rho}_l(t)$:
\begin{equation}\label{eq8}
    p(\bm{\rho}_g|\mathcal{Z}_g(t))=p(\bm{\rho}_g|\bm{\rho}_g(t-1),\bm{\rho}_l(t)),
\end{equation}
precisely, the GDP can be updated as follow:
\begin{equation}\label{eq9}
\setlength{\abovedisplayskip}{3pt}
\setlength{\belowdisplayskip}{3pt}
\scalebox{0.9}{$ \displaystyle
    \begin{aligned}
        &\bm{\rho}_{g}(t)\sim\mathcal{N}(\hat{\bm{\mu}}_g(t), \hat{\bm{\sigma}}^2_g(t)\bm{I}),\\
        &\frac{1}{\hat{\bm{\sigma}}^2_g(t)} = \frac{1}{\hat{\bm{\sigma}}_g^2(t-1)} + \frac{1}{\hat{\bm{\sigma}}^2_l(t)},\\
        &\hat{\bm{\mu}}_g(t)=\hat{\bm{\sigma}}^2_g(t)\Biggl(\frac{\hat{\bm{\mu}}_g(t-1)}{\hat{\bm{\sigma}}_g^2(t-1)}+\frac{\hat{\bm{\mu}}_l(t)}{\hat{\bm{\sigma}}^2_l(t)}\Biggl).\\
    \end{aligned}
    $}
\end{equation}
The GDP has the following properties:
\begin{enumerate}
    \item GDP considers all historical representations, which contribute to GDP according to their probability $\bm{\sigma}^2$.
    \item In prototype calculation, GDP $\bm{\rho}_g(t-1)$ is equivalent to all historical representations $\mathcal{Z}_g(t-1)$ in previous iterations.
\end{enumerate}
The first property of GDP implies that it possesses robustness against instantaneous noisy pseudo-labels if $\bm{\sigma}^2$ is estimated effectively.
And the second property suggests that utilizing GDP to bridge iterations incurs minimal memory cost since current GDP $\bm{\rho}_g(t)$ can be derived from the last GDP $\bm{\rho}_g(t-1)$ and the current local prototype $\bm{\rho}_l(t)$.
Therefore, it is unnecessary to store all previous representations in memory; only the last GDP $\mathcal{Z}_g(t-1)$ needs to be stored.

\subsection{Virtual Negatives}\label{vn_sec}
In order to compensate for the fragmentary distribution of negatives, instead of using the memory bank strategy, we propose an \textit{efficient} strategy, which takes advantage of the distribution of GDP.
We generate Virtual Negatives (VN) from GDP $\bm{\rho}_{(c)g}(t)\sim\mathcal{N}(\hat{\bm{\mu}}_{(c)g}(t), \hat{\bm{\sigma}}^2_{(c)g}(t)\bm{I})$ corresponding to class $c$ via a modified reparameter trick \cite{VAE}:
\begin{equation}\label{eq10}
    \bm{z}^{VN}_c = \hat{\bm{\mu}}_{(c)g}(t) + \beta \bm{\epsilon}^{\top}\bm{I}\hat{\bm{\sigma}}^2_{(c)g}(t),
\end{equation}
where $\bm{\epsilon}=(\epsilon^{(1)},\ldots, \epsilon^{(d)})$, $\epsilon^{(1)},\ldots, \epsilon^{(d)}\sim\mathcal{N}(0,1)$ and $\beta$ is a hyper-parameter we define to balance the compactness and the diversity of VNs, named \textit{virtual radius}.
The reparameter trick essentially samples some representations centred at the $\mu^{(d)}$, taking the Gaussian function $\mathcal{N}(\mu^{(d)},\sigma^{2(d)})$ as the probability density function, and taking $\beta$ as radius for each dimension $d$.
VNs inherit the global features from GDP which covers the entire iterations.
Moreover, VNs exhibit improved compactness compared to real representations, as they are not affected by intra-class variance.
Additionally, VNs offer enhanced dispersion compared to GDP.
In the current iteration, the limited size of the mini-batch restricts the coverage of real negative representations to only a subset of classes in the dataset.
Consequently, this results in a fragmented negative distribution.
However, our VNs have the advantage of encompassing representations from all classes in the dataset.
As a result, they compensate for the fragmented distribution of negative representations and incorporate a broader range of global features.

\noindent\textbf{Discussion about Memory Bank.}
The memory bank strategy, as employed in various studies \cite{RegionContrast,CLCMB,Cipc,U2PL}, has been a conventional solution for bridging iterations and compensating for the fragmentary distribution of negatives.
Originally designed to extend the coverage of images with a limited mini-batch size, the memory bank relies on an elaborate sampling strategy to enqueue and dequeue representations, which can be computationally expensive ($31\%$ slower processing) and memory-intensive ($2.63$ GB).
This is because, compared with instance-level representations, pixel-wise representations are dense, and each pixel (or region) in the image is mapped to a corresponding representation.
To overcome these limitations, we propose a new strategy that leverages GDP and VNs to complement the fragmentary distribution of negatives without the need for an intricate sampling strategy to enqueue and dequeue representations.
Particularly, our method aggregates global information into GDP using only \textbf{$\bm{42}$KB (v.s. $\bm{2.63}$ GB)} of memory and generates VNs with minimal computational cost (increase training time by \textbf{$\bm{0.03}$ v.s. $\bm{0.47}$ GPU days)}.
By using this approach, we can capture a larger amount of image information at scale without the significant memory and computational overhead associated with the memory bank strategy.
The experimental proof can be found in Sec.~\ref{ab_vn}

\subsection{Training Objective}
Cooperated with the conventional teacher-student framework and our introduced components, the total training object is composed of a supervised loss $\mathcal{L}_s$, an unsupervised loss $\mathcal{L}_u$, and a contrastive loss $\mathcal{L}_c$ as follows:
\begin{equation}\label{eq11}
\scalebox{0.9}{$ \displaystyle
    \mathcal{L} = \mathcal{L}_{s} + \mathcal{L}_{u} + \lambda_c(t)\mathcal{L}_{c}, 
    $}
\end{equation}
where $\lambda_c(t)$ is used to tune the contribution of contrastive loss and formulated as:
\begin{equation}\label{eq12}
    \scalebox{0.9}{$ \displaystyle
    \lambda_c(t)=\lambda_{c0}\cdot exp\Biggl(\alpha \cdot (\frac{t}{T_{total}})^2\Biggl),
    $}
\end{equation}
where $\lambda_{c0}$ denotes the initial scaling parameter, $\alpha$ denotes a weight decay coefficient, $t$ denotes the current $t^{th}$ epoch and $T_{total}$ denotes the total epochs.

$\mathcal{L}_{s}$ is constructed by standard Cross-Entropy (CE) loss $\ell_{ce}$ and formulated as:
\begin{equation}\label{eq13}
\scalebox{0.9}{$ \displaystyle
    \mathcal{L}_s = \frac{1}{|\mathcal{B}_l|}\sum_{(\bm{x}_i^l, \bm{y}_i^l)\in\mathcal{B}_l}\ell_{ce}(g(f(\bm{x}_i^l), \bm{y}_i^l)),
    $}
\end{equation}
where $\mathcal{B}_l$ represent the batch of labeled images.

For $\mathcal{L}_{u}$, we first set a threshold $\delta_u$ to count the number $\hat{N}$ of training pixels whose corresponding confidence $p^u_i$ is higher than $\delta_u$ in the training set $\mathcal{B}_u$.
With $\hat{N}$ and the total number $N$ of the pixels in $\mathcal{B}_u$, the $\mathcal{L}_{u}$ is constructed by the weighted CE loss, formulated as:
\begin{equation}\label{eq14}
\scalebox{0.9}{$ \displaystyle
    \mathcal{L}_u = \frac{1}{|\mathcal{B}_u|}\sum_{\bm{x}_i^u\in\mathcal{B}_u}\omega\ell_{ce}(g(f(\bm{x}_i^u), \bm{y}_i^u)),
    $}
\end{equation}
where $\bm{y}_i^u$ is the pseudo-labels from teacher model, and $\omega$ is the loss weight, formulated by $\omega=\frac{\hat{N}}{N}$.

While calculating contrastive loss, since the limited GPU memory, we sample valuable representations for the contrast, following prior works \cite{Reco}.
We adopt some sampling strategies according to confidence $p_i$ and introduce strong threshold $\delta_s$ and weak threshold $\delta_w$ in these strategies.
Our sampling strategies are as follows:
\textbf{1) Valid representations sampling strategy}:
We set $\delta_w$ for sampling valid representations whose $p_i$ is higher than $\delta_w$.
Only valid representations will be considered in the contrast.
\textbf{2) Anchors sampling strategy}:
We set $\delta_s$ for sampling anchors whose corresponding $p_i$ is lower than $\delta_s$.
\textbf{3) Negatives sampling strategy}:
We non-uniformly sample negatives in different classes based on the similarity (i.e., MLS \cite{PFE}) between GDPs of negative classes and the GDP of the current anchor class.

We apply InfoNCE \cite{CPC} as our contrastive loss and introduce the PR, GDP, and VN (describe in Sec.~\ref{pr}, Sec.~\ref{gdp}, and Sec.~\ref{vn_sec}, respectively) to it, formulated as :
\begin{equation}\label{eq15}
\scalebox{0.65}{$ \displaystyle
    \begin{aligned}
        &\mathcal{L}_{c}= -\frac{1}{|C| \times |\mathcal{Z}_c|} \sum _{c\in C} \sum_{\bm{z}_{c i} \in \mathcal{\bm{Z}}_c}\\
        &\log[\frac{e^{s(\bm{z}_{ci},\bm{\rho}_{(c)g}(t))/ \tau}}
        {e^{s(\bm{z}_{ci}, \bm{\rho}_{(c)g}(t))/ \tau}+\sum_{\tilde{c} \in \tilde{C}_{l}}\sum_{\bm{z}_{\tilde{c} j} \in \mathcal{Z}_{\tilde{c}}}e^{s(\bm{z}_{c i},\bm{z}_{\tilde{c} j})/ \tau} + \sum_{\tilde{c} \in \tilde{C}_{g}}\sum_{\bm{z}^{VN}_{\tilde{c} j} \in \mathcal{Z}^{VN}_{\tilde{c}}}e^{s(\bm{z}_{c i},\bm{z}^{VN}_{\tilde{c} j})/ \tau}}],
    \end{aligned}
    $}
\end{equation}
where $C$ represents the set of all anchor classes in the current iteration, $\mathcal{Z}_c$ represents the set of anchor representations $\bm{z}_{ci}$ belonging to class $c$, $\tilde{C}_{l}$ represents the negative classes in the current iteration, $\tilde{C}_{g}$ represents negative classes in the entire dataset, $\mathcal{Z}_{\tilde{c}}$ represents the set of real negative representations $\bm{z}_{\tilde{c} j}$ in the current iteration, $\mathcal{Z}_{\tilde{c}}^{VN}$ represents the set of virtual negatives $\bm{z}_{\tilde{c}j}^{VN}$ belonging to class $c$, $\bm{\rho}_{(\tilde{c})g}(t)$ represents the current GDP belonging to class $c$, $\tau$ represents the temperature parameter, $s$ is MLS.
And we pad zero to the probability of VNs empirically, due to the probability missing in the modified reparameter trick.

For training our probabilistic head, we adopt a strategy called soft freeze, following [1]. Specifically, we separate the training of the probability head from the training of the backbone and segmentation head and endow the probability head with a small learning rate. The reason for adopting this strategy is to guarantee the stability of the training process. At the beginning of training, the outputs of probability head climb dramatically since the representations are unreasonable and meaningless at that time. Therefore, it is essential to endow the probability head with a much smaller learning rate than the backbone so that its training process is able to keep pace with others and the different modules in the network can interact with each other.
Our framework is in Fig.~\ref{framework}.
Meanwhile, the process of training is demonstrated in Algorithm~\ref{algorithm}.
\begin{algorithm*}[t]
\caption{Pseudo-code of the training process in a Pytorch-like style}
\label{algorithm}
\textbf{Network}: Student encoder: $f$, student segmentation head: $g$, teacher encoder: $f'$, teacher segmentation head: $g'$, representation head: $h$, probability head: $p$\\
\textbf{Input}: Mini-batch $B$ consists of $(X^l, Y^l)$ and $(X^u)$, last GDP $\bm{\rho}_g(t-1)$\\
\textbf{Notation}: Anchor class $c$, remaining classes in current iteration $\tilde{C}_l$, remaining classes in dataset $\tilde{C}_g$\\
\begin{algorithmic}[1] 
\FOR{$epoch$ in range($total\_epoch$)}
\STATE $P^l=g(f(X^l))$ \hfill\textcolor{green!40!black!50}{\# Predict on labeled images}
\STATE $\mathcal{L}_s=ce\_loss(P^l, Y^l)$ \hfill\textcolor{green!40!black!50}{\# Calculate supervised loss $\mathcal{L}_s$}
\STATE $Y^u=max\_op(g'(f'(X^u)))$ \hfill\textcolor{green!40!black!50}{\# Generate pseudo-labels via max operation in Eq.~\ref{eq1}}
\STATE $P^u=g(f(X^u))$ \hfill\textcolor{green!40!black!50}{\# Predict on unlabeled images}
\STATE $\mathcal{L}_u=ce\_loss(P^u, Y^u)$ \hfill\textcolor{green!40!black!50}{\# Calculate unsupervised loss $\mathcal{L}_u$}
\STATE $(X, Y)\leftarrow combine((X^l, Y^l), (X^u, Y^u))$\hfill\textcolor{green!40!black!50}{\# Combine labeled and unlabeled training set}
\FOR{$(X, Y) \in B$}
\STATE $\mathcal{L}_{c}$ = 0 \hfill\textcolor{green!40!black!50}{\# Initialize $\mathcal{L}_{c}$}
\STATE $\mu=h(f(X))$ \hfill \textcolor{green!40!black!50}{\# Calculate  $\bm{\mu}$}
\STATE $\sigma^2=p(f(X))$ \hfill \textcolor{green!40!black!50}{\# Calculate  $\bm{\sigma}^2$}
\STATE $Z \leftarrow (\mu, \sigma^2)$ \hfill \textcolor{green!40!black!50}{\# Representations consisting of $\bm{\mu}$ and $\bm{\sigma}^2$}
\STATE $Z_{val}, Y_{val}, C_{val} \leftarrow mask(Z, Y)$ \hfill \textcolor{green!40!black!50}{\# Mask with $\delta_w$ according to sampling strategy}
\FOR{$c \in C_{val}$}
\STATE $\bm{\rho}_l^c(t) \leftarrow calculate\_prt(Z_{val}^c)$ \hfill \textcolor{green!40!black!50}{\# Calculate local prototype with Eq.~\ref{eq4}}
\STATE $\bm{\rho}_g^c(t) \leftarrow update\_prt(\bm{\rho}_g^c(t-1), \bm{\rho}_l^c(t))$ \hfill\textcolor{green!40!black!50}{\# Update GDP with Eq.~\ref{eq9}}
\ENDFOR
\FOR{$c \in C_{val}$}
    \STATE $\mathcal{L}$ = 0 \hfill\textcolor{green!40!black!50}{\# Initialize loss $\mathcal{L}$ in current class}
\STATE $Z_{a}^c \leftarrow sample\_a(Z_{val}^c, Y_{val}^c)$ \hfill\textcolor{green!40!black!50}{\# Sample anchor representations}
\STATE $neg\_dist \leftarrow [MLS(\bm{\rho}_g^c(t), \bm{\rho}_g^{\tilde{c}_l}(t))~for~\tilde{c}_l~in~\tilde{C}_l]$ \hfill \textcolor{green!40!black!50}{\# Calculate negative sampling distribution}
\STATE $Z_{n} \leftarrow sample\_n(Z_{val}^{\tilde{c}_l}, neg\_dist)$ \hfill\textcolor{green!40!black!50}{\# Sample real negative representations}
\STATE $Z_{VN} \leftarrow [generate\_VN(\bm{\rho}_g^{\tilde{c}_g}(t))~for~\tilde{c}_g~in~\tilde{C}_g]$ \hfill \textcolor{green!40!black!50}{\# Generate virtual negatives with Eq.~\ref{eq10}}
\STATE $\mathcal{L} = contrast\_loss(Z_{a}^c, Z_{n}, Z_{VN}, \bm{\rho}_g^c(t))$ \hfill\textcolor{green!40!black!50}{\# Calculate $\mathcal{L}_{c}$ with Eq.~\ref{eq15}}
\STATE $\mathcal{L}_{c} = \mathcal{L}_{c} + \mathcal{L}$
\ENDFOR
\ENDFOR
\STATE $\mathcal{L}_{total}=\mathcal{L}_s+\mathcal{L}_u+\lambda\mathcal{L}_c$ \hfill\textcolor{green!40!black!50}{\# Calculate the total loss $\mathcal{L}_{total}$, $\lambda$ is the current value of $\lambda_c(t)$ in Eq.~\ref{eq11}}
\STATE optimizer.zero\_grad()
\STATE $\mathcal{L}_{total}$.backward()
\STATE optimizer.step()
\ENDFOR
\end{algorithmic}
\end{algorithm*}
\section{Experiments}\label{exp}
\subsection{Setup}
\noindent\textbf{Datasets.}
We conduct experiments on PASCAL VOC 2012 dataset \cite{pascal} and Cityscapes dataset \cite{cityscapes} to validate the effectiveness of our framework.
The PASCAL VOC 2012 contains 1464 well-annotated images in \texttt{train} set and 1449 images in \texttt{val} set originally.
We include 9118 images from SBD \cite{SBD} as additional training images.
Since the SBD dataset is coarsely annotated, we use both \textit{classic} VOC \texttt{train} set (1464 candidate labeled images) and \textit{blender} VOC \texttt{train} set (10582 candidate labeled images), following \cite{U2PL}.
Cityscapes contains 2975 images in \texttt{train} set and 500 images in \texttt{val} set.

\noindent\textbf{Network structure.}
We choose Deeplabv3+ \cite{deeplabv3+} as our structure with ResNet \cite{resnet} pre-trained on ImageNet \cite{imagenet} as the backbone. The segmentation and representation heads are composed of \texttt{Conv-BN-ReLU-Conv}.
The probability head is composed of \texttt{Linear-BN-ReLU-Linear-BN}.

\noindent\textbf{Implementation details.}
For both two datasets, we use stochastic gradient descent (SGD) optimizer.
For training on PASCAL VOC 2012, the initial learning rate is $6.4 \times 10^{-3}$ for the backbone, segmentation head, and representation head while $5 \times 10^{-5}$ for the probability head.
For training on Cityscapes, the initial learning rate is $6.4 \times 10^{-3}$ for the backbone, segmentation head, and representation head while $5 \times 10^{-5}$ for the probability head.
We use the poly scheduling to decay the learning rate during training: $lr = lr_{base}(1-\frac{iter}{total_{iter}})^{0.9}$.
For PASCAL VOC 2012, we set the image crop size as 512 $\times$ 512, and the batch size as 16.
For Cityscapes, we set the image crop size as 768 $\times$ 768, and the batch size as 16.
The models are trained for 80,000 and 160,000 iterations on PASCAL VOC 2012 and Cityscapes when compared with SOTAs, respectively.
Exceptionally, for the ablation study, we train models for 40,000 iterations on PASCAL VOC 2012 and the batch size is set to 8.

\noindent\textbf{Evaluation metric.}
We use mean intersection-over-union (mIoU) as our metric for evaluation.
Meanwhile, following \cite{CPS}, we employ the slide window strategy to evaluate the performance of the Cityscapes dataset.

\subsection{Comparing with Existing Methods}
In this subsection, we conduct a comprehensive evaluation of our proposed method by comparing it against several baselines and state-of-the-art (SOTA) approaches.
Specifically, we reproduce Mean Teacher (MT) \cite{MT} and ClassMix \cite{classmix} on \textit{classic} VOC \texttt{train} set and Cityscapes \texttt{train} set.
It is worth noting that it is not hard to apply our components to most contrastive-based S4 works \cite{Reco,U2PL,CLCMB,Cipc,density_guided}.
For simplicity, we use a contrastive-based teacher-student framework as our \textbf{Baseline}.
The baseline framework has the same architecture as our framework but without PR, GDP, prototype update strategy, and VN.
In addition, we design \textbf{Baseline+} which adds the prototype update strategy.
The prototype in \textbf{Baseline+} is updated by EMA, which is formulated by 
\begin{equation}\label{eq16}
    \scalebox{0.9}{$ \displaystyle
    \bm{\rho}_g^{EMA}(t) = \alpha \bm{\rho}_g^{EMA}(t-1) + (1-\alpha) \bm{\rho}_l(t),
    $}
\end{equation}
where $\bm{\rho}_g^{EMA}(t)$ denotes the updated prototype, $\bm{\rho}_g^{EMA}(t-1)$ denotes the previous prototype, $\bm{\rho}^{EMA}_l(t)$ denotes the local prototype from the current mini-match, and $\alpha$ is hyper-parameter to control the update speed, set as $0.99$.

Meanwhile, we compare our method on \textit{blender} VOC \texttt{train} set with following recent SOTA S4 methods: CCT \cite{CCT}, CPS \cite{CPS}, U$^2$PL \cite{U2PL}, ST++ \cite{ST++}, PSMT \cite{PSMT}, ELN \cite{eln}.
Besides, we compare our method on Cityscapes \texttt{train} set with the following methods: CCT \cite{CCT}, CPS \cite{CPS}, U$^2$PL \cite{U2PL}, and PSMT \cite{PSMT}.
For a fair comparison, following \cite{U2PL,CPS}, we use a modified ResNet-101 with the stem block to compare with SOTAs while using the original ResNet-101 to compare with baselines in both two datasets.

\noindent\textbf{Results on PASCAL VOC 2012.} Tab.~\ref{tab1} shows the results on \textit{classic} VOC \texttt{train} set, comparing with baselines.
Our method consistently outperforms baselines at all label rates.
Tab.~\ref{tab2} compares our method with the SOTA methods, our method shows the competitive performance in a wide range of label rates.
Since our contribution lies in contrastive learning derived from self-training, the performance is more advantageous in the case of the few labels.

\noindent\textbf{Results on Cityscapes.}
Tab.~\ref{tab3} presents the comparison results on the Cityscapes dataset. Our proposed method exhibits a slight but consistent improvement in performance compared to all baseline methods and SOTA approaches. This improvement is observed across various label rates, indicating the robustness and effectiveness of our approach in different scenarios.
\begin{table}[t]
\centering
\caption{Results on \textit{classic} VOC \texttt{train} set using original ResNet-101. All approaches are reproduced. Labeled data is from the original VOC \texttt{train} set.}
\setlength{\tabcolsep}{1.4mm}{%
\begin{tabular}{c|ccccc}
\hline
\multicolumn{6}{c}{PASCAL VOC 2012 (\textit{Classic})}                                           \\ \hline
Method                             & 92          & 183         & 366         & 732         & 1464    \\ \hline
Supervised                         & $52.38$     & $55.32$     & $65.92$     & $71.59$     & $72.90$       \\
MT \cite{MT}                       & $46.84$     & $61.35$     & $66.86$     & $71.93$     & $74.00$      \\
ClassMix \cite{classmix}           & $63.83$     & $67.20$     & $71.23$     & $73.98$     & $76.91$       \\
Baseline                           & $66.91$     & $69.73$     & $72.01$     & $74.99$     & $76.85$       \\
Baseline+                           & $68.27$     & $70.32$     & $73.97$     & $75.61$     & $77.21$       \\ \hline
PRCL                                & $\bm{70.23}$     & $\bm{72.20}$     & $\bm{75.17}$     & $\bm{76.24}$     & $\bm{78.29}$       \\ \hline

\end{tabular}
}
\label{tab1}
\end{table}
\begin{table}[t]
\centering
\caption{Results on \textit{blender} VOC \texttt{train} set using a modified ResNet-101. All the results from the recent papers \cite{U2PL,ST++,eln,PSMT}. Labeled data is from the augmented VOC \texttt{train} set.}. 
\setlength{\tabcolsep}{1.2mm}{%
\begin{tabular}{c|c|cccc}
\hline
\multicolumn{6}{c}{PASCAL VOC 2012 (\textit{Blender})}                             \\ \hline
Method                 & Publication & 662          & 1323         & 2646         & 5291 \\ \hline
CCT \cite{CCT}         & CVPR 20     & $71.86$      & $73.68$      & $76.51$      & $77.40$      \\
CPS \cite{CPS}         & CVPR 21     & $74.48$      & $76.44$      & $77.68$      & $78.64$      \\
U$^2$PL \cite{U2PL}    & CVPR 22     & $77.21$      & $79.01$      & $79.30$      & $\bm{80.50}$      \\
ST++ \cite{ST++}       & CVPR 22     & $74.70$      & $77.90$      & $77.90$      & -          \\
PSMT \cite{PSMT}       & CVPR 22     & $75.50$      & $78.20$      & $78.72$      & $79.76$      \\
ELN \cite{eln}         & CVPR 22     & -            & $75.10$      & $76.58$      & -          \\ \hline
PRCL                    & -           & $\bm{77.87}$ & $\bm{79.09}$ & $\bm{79.85}$ & $\underline{80.11}$     \\ \hline
\end{tabular} %
}
\label{tab2}
\end{table}
\begin{table}[t]
\centering
\caption{Results on Cityscapes. The model is trained on the Cityscapes \texttt{train} set, which consists of 2975 samples in total, and tested on the Cityscapes \texttt{val} set. And all the results from the recent papers \cite{U2PL,PSMT}. $\dagger$ means we reproduce the approach.}
\setlength{\tabcolsep}{0.68mm}{%
\begin{tabular}{l|c|cccc}
\hline
\multicolumn{6}{c}{Cityscapes}                                                               \\ \hline
Method                             & Publication     & 186          & 372         & 744         & 1488   \\ \hline
Supervised$^\dagger$               & -              & $62.04$      & $65.71$     & $69.48$     & $70.06$      \\
MT$^\dagger$ \cite{MT}             & NeurIPS 17     & $67.06$      & $68.43$     & $70.05$     & $70.64$      \\
ClassMix$^\dagger$ \cite{classmix} & WACV 21        & $67.98$      & $69.58$     & $72.21$     & $72.82$      \\
CCT \cite{CCT}         & CVPR 20     & $69.32$      & $74.12$      & $75.99$      & $77.40$      \\
CPS \cite{CPS}         & CVPR 21     & $69.78$      & $74.31$      & $74.58$      & $76.82$      \\
U$^2$PL \cite{U2PL}    & CVPR 22     & $70.30$      & $74.37$      & $76.47$      & $79.05$      \\
PSMT \cite{PSMT}       & CVPR 22     & -            & $76.89$      & $77.60$      & $79.09$      \\  \hline
PRCL                                & -             & $\bm{73.38}$ & $\bm{77.08}$& $\bm{77.89}$& $\bm{79.95}$      \\ \hline
\end{tabular} %
}
\label{tab3}
\end{table}

\subsection{Other Results}
\begin{figure*}[t]
  \centering
  \includegraphics[width=0.8\linewidth]{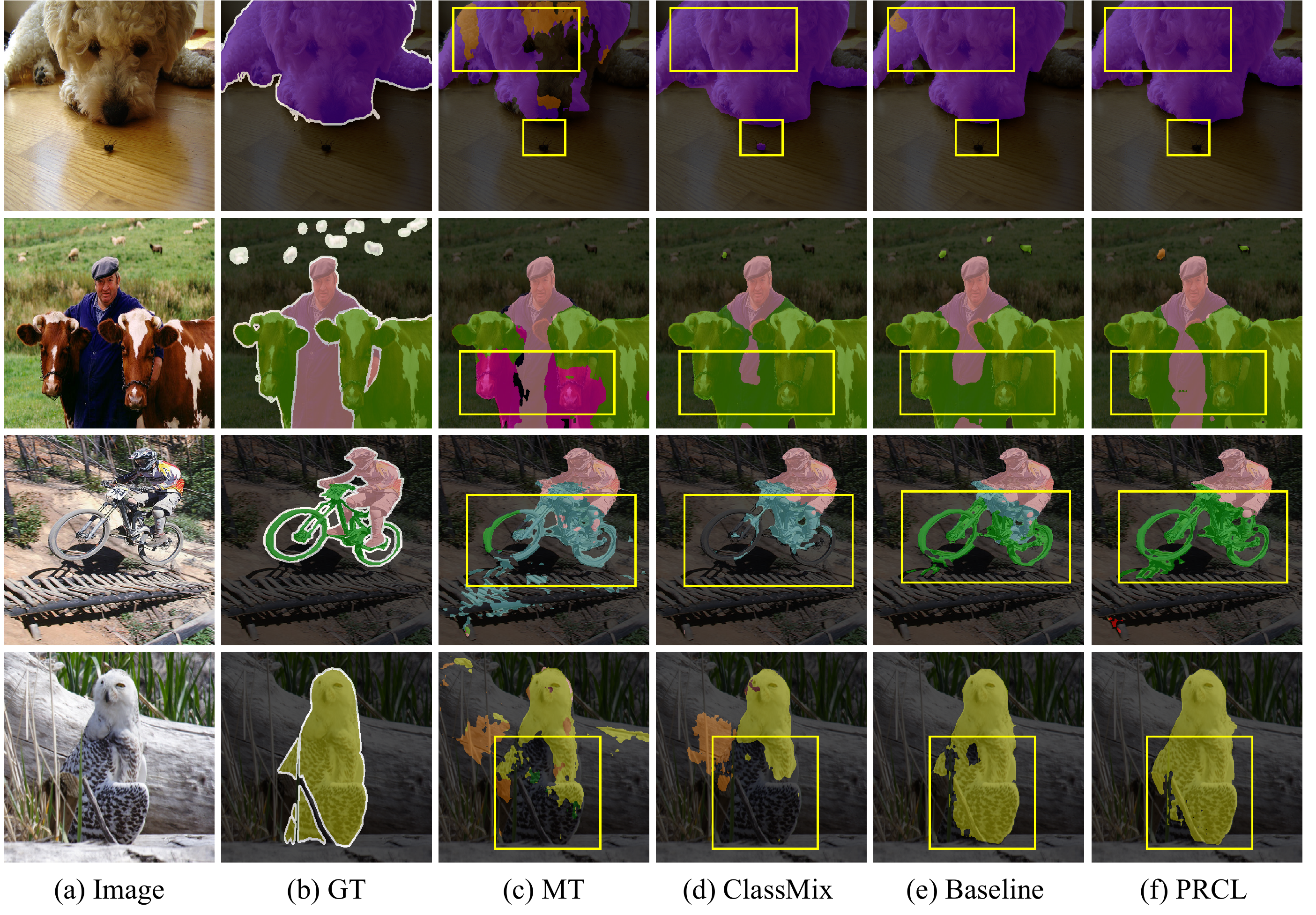}
  \caption{Visualisation on PASCAL VOC 2012. All models are trained with 92 labeled images. The differences are highlighted in yellow boxes.
   }
  \label{vis_voc}
\end{figure*}

\begin{figure*}[t]
  \centering
  \includegraphics[width=1.0\linewidth]{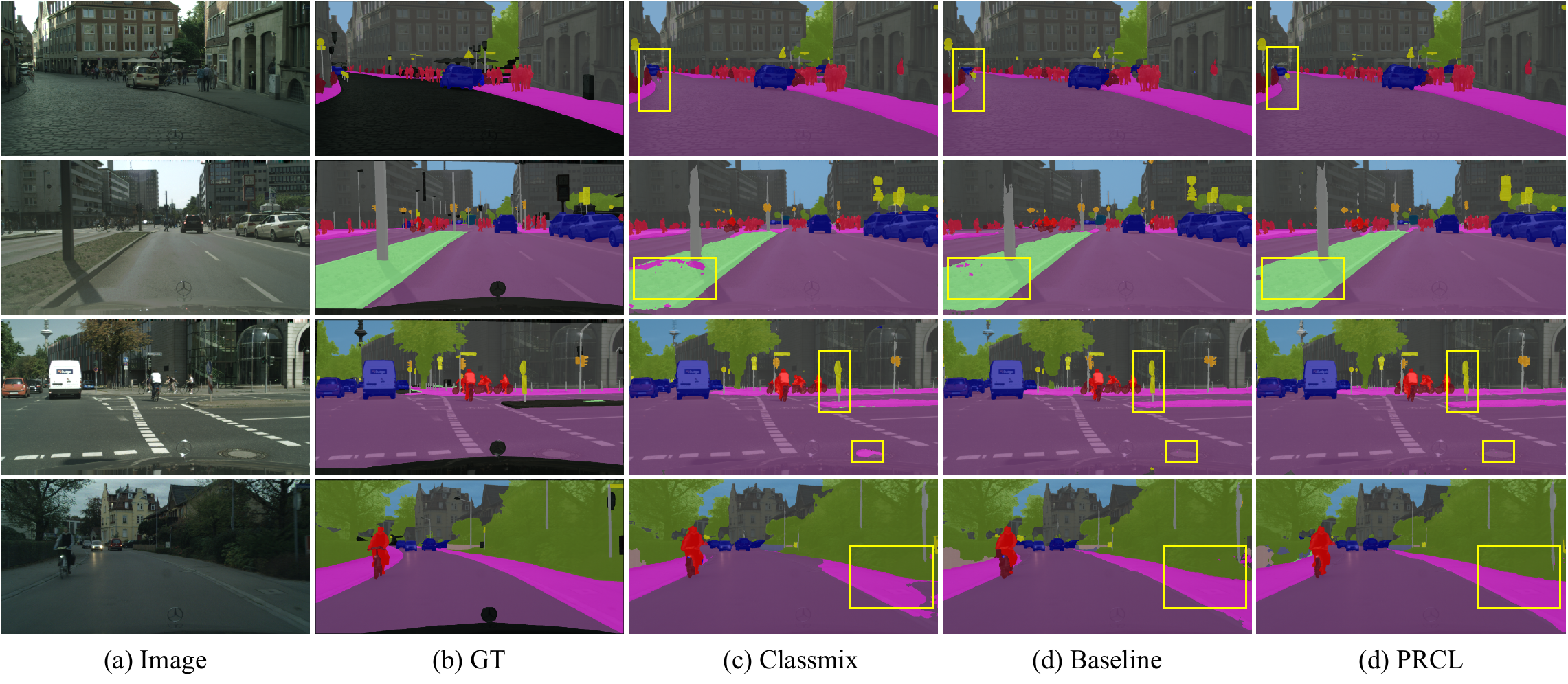}
  \caption{Visualisation on Cityscapes. All models are trained with 186 labeled images. The differences are highlighted in yellow boxes.
   }
  \label{vis_city}
\end{figure*}

The qualitative results of different methods on the PASCAL VOC 2012 are shown in Fig.~\ref{vis_voc}.
Our PRCL framework demonstrates superior performance compared to the other methods, benefiting from its enhanced robustness.
While our baseline shows some improvement over the self-training framework with MT and Classmix, it still exhibits limitations in handling ambiguous regions, such as the boundaries between different classes.
In contrast, our PRCL framework performs better in these ambiguous regions, visually showcasing the effectiveness of our approach.
Fig.~\ref{vis_city} shows the qualitative results on Cityscapes, further validating the superiority of our framework.

\begin{figure}[t]
  \centering
  \includegraphics[width=1.0\linewidth]{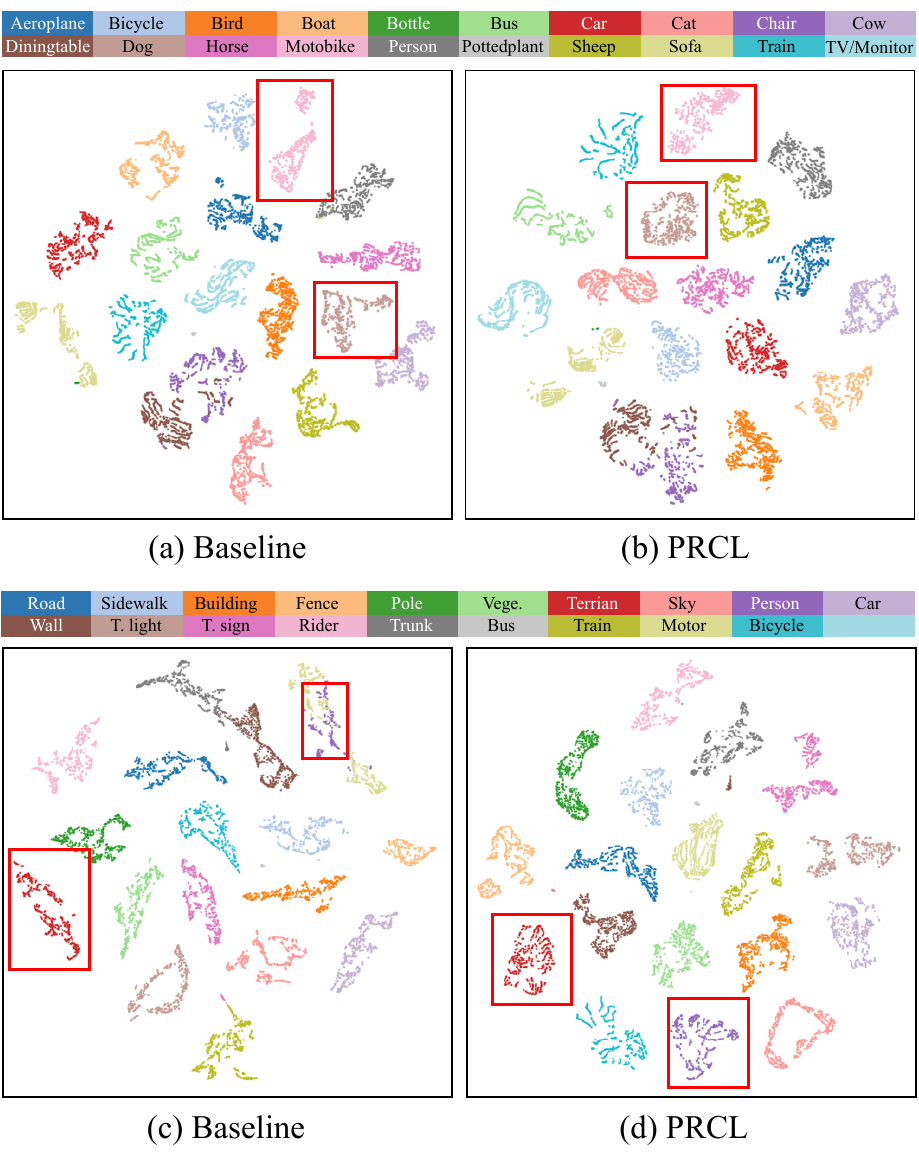}
  \caption{Representation visualizations. The differences are highlighted in red boxes.
   }
  \label{vis_tsne}
\end{figure}

Additionally, to gain a more intuitive understanding of the advantages offered by our framework, we provide a quantitative comparison in representation space between both the baseline and our framework and two t-SNE plots to visualize the distribution of representations in the latent space for both the baseline and our framework.
Tab.~\ref{tab_compact} shows the results of the quantitative comparison. We use the Silhouette score \cite{silhouettes} (Silhouette) and Davies–Bouldin Index \cite{DBI} (DBI) as metrics.
These two measurements are both able to show the cohesion of a representation and its cluster and the separation of a representation and other clusters. As for the Silhouette Score, a larger value indicates better performance. In contrast, the smaller value of the Davies-Bouldin Index stands for better performance.
The results show that our framework performs better than our baseline in representation space in two different datasets.
Fig.~\ref{vis_tsne} \textcolor{red}{(a)} and \textcolor{red}{(b)} show that our method results in a better representation distribution compared to the baseline on PASCAL VOC 2012.
We can observe that our representation is more compact than the baseline, especially the region highlighted in red boxes.
This is because our framework is more robust to inaccurate pseudo-labels and provides a consistent direction for representation aggregation.
Fig.~\ref{vis_tsne} \textcolor{red}{(c)} and \textcolor{red}{(d)} show the results on Cityscapes, which further prove the effectiveness of our framework.

\begin{table}[h]
\centering
\caption{Performance of two frameworks in representation space.}
\setlength{\tabcolsep}{2.6mm}{%
\begin{tabular}{c|cc|cc}
\hline
Dataset  & \multicolumn{2}{c|}{Pascal VOC 2012} & \multicolumn{2}{c}{Cityscapes} \\ \hline
Metrics  & Silhouette          & DBI            & Silhouette       & DBI         \\ \hline
Baseline & $0.3188$              & $1.1221$         & $0.2708$           & $1.9472$      \\
PRCL     & $0.3836$              & $1.0334$         & $0.3014$           & $1.7667$      \\ \hline
\end{tabular}
}
\label{tab_compact}
\end{table}

\section{Ablative Study}\label{ablative}
The main contribution of our work lies in \textbf{1)} probabilistic representation, \textbf{2)} global distribution prototype, and \textbf{3)} virtual negatives.
We conduct experiments to further explore the effectiveness and rationality of our components. 
We choose Deeplabv3+ with ResNet-101 pre-trained on ImageNet as our backbone and PASCAL VOC 2012 as our dataset.
The baseline and baseline+ framework and other settings are the same as those in Sec.~\ref{exp}.

\begin{figure*}[t]
  \centering
  \includegraphics[width=1.0\linewidth]{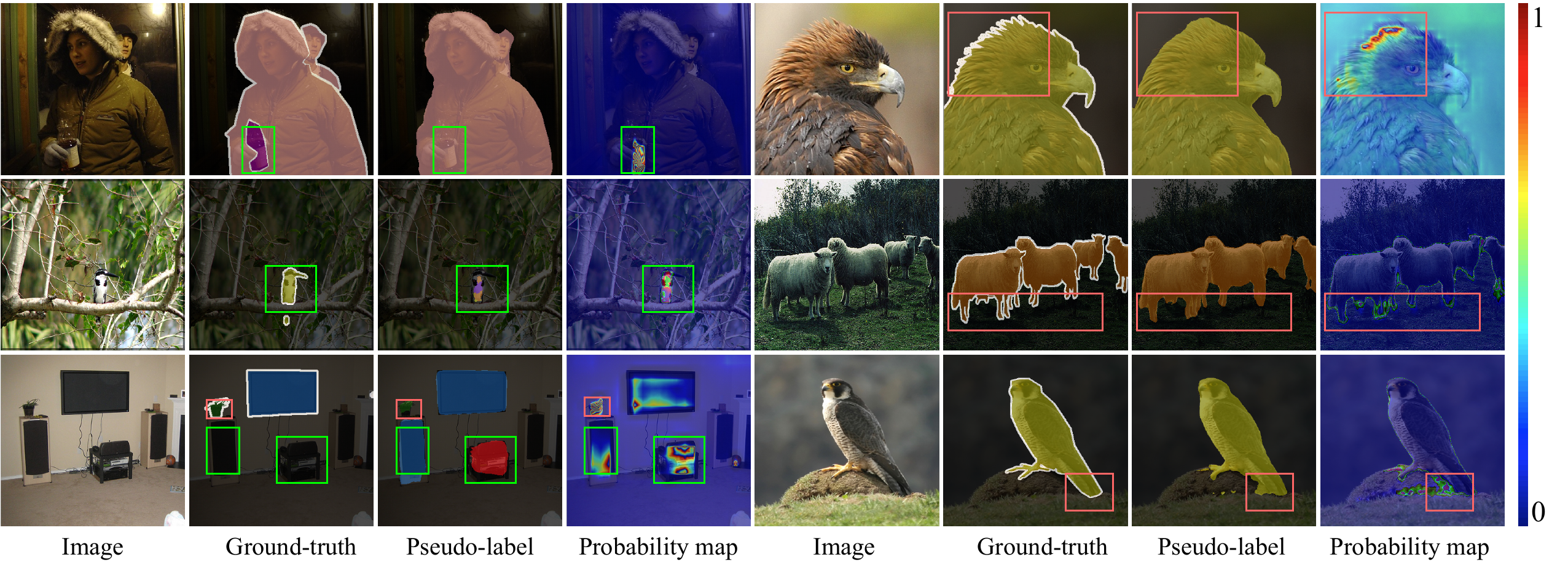}
  \caption{Visualization of probability behavior.
   }
  \label{vis_of_prob}
\end{figure*}

\subsection{Effect of Probabilistic Representation}
\noindent\textbf{Behaviors of the probability.}
To visualize the relationship between probability and inaccurate pseudo-labels, we provide visualizations of the model's predictions along with their corresponding probabilities.
In addition, we apply $\ell_1$ normalization to $\bm{\sigma}^2$ for visualization purposes.
In Fig.~\ref{vis_of_prob}, columns from left to right represent input image, ground-truth, pseudo-label, and probability map, respectively. For the probability map, the red color represents the large $\bm{\sigma}^2$ (indicating low probability).
In the visualizations, we use green boxes to highlight the mismatches caused by inaccurate pseudo-labels, such as instances mistakenly labeled as a person or a bottle.
On the other hand, red boxes are used to mark fuzzy pixels, such as the furry edge of a bird.
These cases are specifically identified by the $\bm{\sigma}^2$ values and are observed to have a relatively low contribution during the contrastive training process.

\noindent\textbf{Results of using probabilistic representation.}
To explore the impact of probabilistic representation, we conduct the experiments as follows: baseline \textbf{1)} without probabilistic representation (w/o PR), and \textbf{2)} with probabilistic representation (w/ PR).
Tab.~\ref{tab4} shows the effectiveness of probabilistic representation across various label rates.
This can be attributed to introducing probability in our representation, which allows us to reduce the negative effect of inaccurate pseudo-labels during contrastive learning.
Consequently, PR endows our framework the greater robustness compared to conventional contrastive-based teacher-student frameworks with deterministic representations.
Meanwhile, our approach can be easily applied to other contrastive-based S4 frameworks by substituting probabilistic representations for deterministic representations.
\begin{table}[h]
\centering
\caption{Results on the effect of the probabilistic representation.}
\setlength{\tabcolsep}{3.4mm}{%
\begin{tabular}{c|cccc}
\hline
\multicolumn{5}{c}{PASCAL VOC 2012}                                           \\ \hline
Label rates                             & 92          & 183         & 366         & 732        \\ \hline
w/o PR                         & $66.91$     & $69.73$     & $72.01$     & $74.99$         \\
w/ PR                          & $68.49$     & $71.79$     & $74.36$     & $76.00$         \\ \hline

\end{tabular}
}
\label{tab4}
\end{table}

\subsection{Effect of Global Distribution Prototype}
To investigate the influence of the prototype update strategy, we conduct experiments \textbf{a)} baseline without update strategy, \textbf{b)} baseline with EMA update strategy (baseline+), \textbf{c)} baseline without update strategy but with probabilistic representation, and \textbf{d)} baseline with update strategy and probabilistic representation.

\begin{figure}[t]
  \centering
  \includegraphics[width=0.8\linewidth]{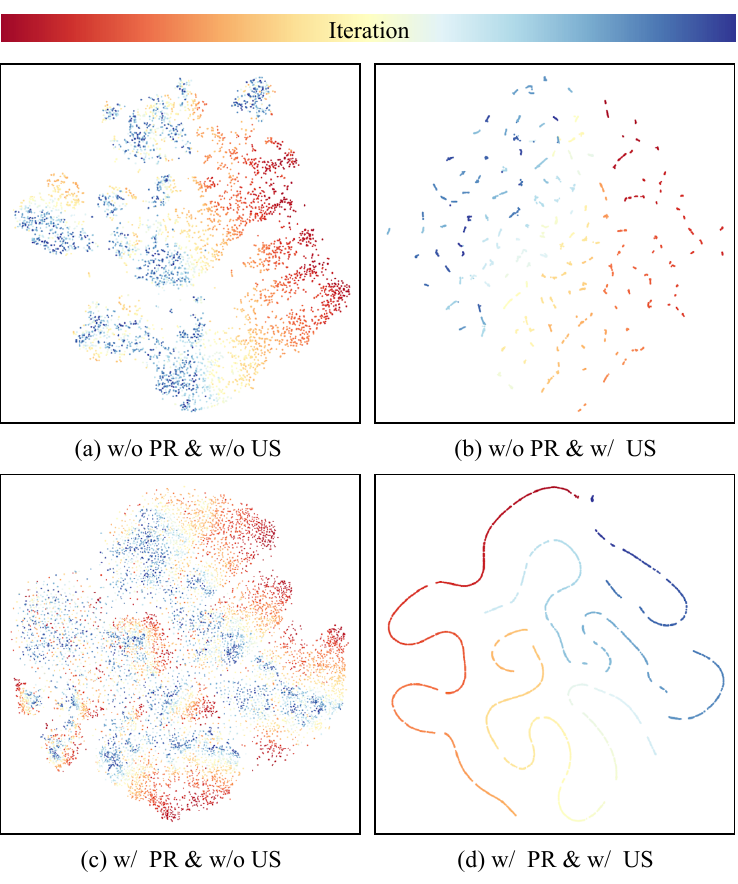}
  \caption{Visualization of prototype behaviour. w/o PR and w/ PR mean without probabilistic representation and with probabilistic representation, respectively. Similarly, w/o US and w/ US mean without prototype update strategy and with prototype update strategy, respectively.
   }
  \label{proto_be}
\end{figure}

\noindent\textbf{Behavior of prototype.}
We visualize the changes in prototypes during the training process in the conducted experiments through t-SNE \cite{t-sne}.
As shown in Fig.~\ref{proto_be} \textcolor{red}{(a)} and Fig.~\ref{proto_be} \textcolor{red}{(c)}, without any strategy, there is a noticeable prototype shift observed between prototypes in two adjacent iterations.
This significant shift in distance leads to inconsistent directions for the aggregation of anchor representations, thereby hindering the effective aggregation of representations.
Although the EMA strategy also updates prototypes based on previous ones, its behavior exhibits instability and discontinuity throughout the training process, as depicted in Fig.~\ref{proto_be} \textcolor{red}{(b)}.
Due to its limited global property, the EMA strategy is sensitive to instant noisy pseudo-labels, also leading to prototype shifts resulting from incorrect representation assignments.
In contrast, our GDP strategy demonstrates greater stability and robustness against instant noisy pseudo-labels.
It benefits from its fully global property, which encompasses all historical information.
As shown in Fig.~\ref{proto_be} \textcolor{red}{(d)}, the GDP strategy maintains stable and continuous behavior throughout the training process.

\noindent\textbf{Results of different update strategies.}
Tab.~\ref{tab5} shows the impact of two different update strategies on two types of representations: EMA (Eq.~\ref{eq16}) update strategy on deterministic representations (Vanilla), and our GDP (Eq.~\ref{eq9}) update strategy on probabilistic representation (PR) on PASCAL VOC 2012 with a wide range of label rate.
Both update strategies contribute to performance improvement by ensuring consistency in the representation space.
However, our GDP approach yields the best outcomes, demonstrating its superiority over the conventional EMA approach.
\begin{table}[h]
\caption{Impact of Update Strategy (US)}
\setlength{\tabcolsep}{3.8mm}{%
\begin{tabular}{c|cccc}
\hline
Vanilla Rep.                             & 92          & 183         & 366         & 732        \\ \hline
w/o US                         & $66.91$     & $69.73$     & $72.01$     & $74.99$         \\
w/ US                          & $68.27$     & $70.32$     & $73.97$     & $75.61$         \\ \hline
PR                             & 92          & 183         & 366         & 732        \\ \hline
w/o US                         & $68.49$     & $71.79$     & $74.36$     & $76.00$         \\
w/ US                          & $69.52$     & $72.20$     & $75.17$     & $76.24$         \\ \hline
\end{tabular}
}
\label{tab5}
\end{table}

\subsection{Effect of Virtual Negatives}\label{ab_vn}
\begin{figure}[t]
  \centering
  \includegraphics[width=1.0\linewidth]{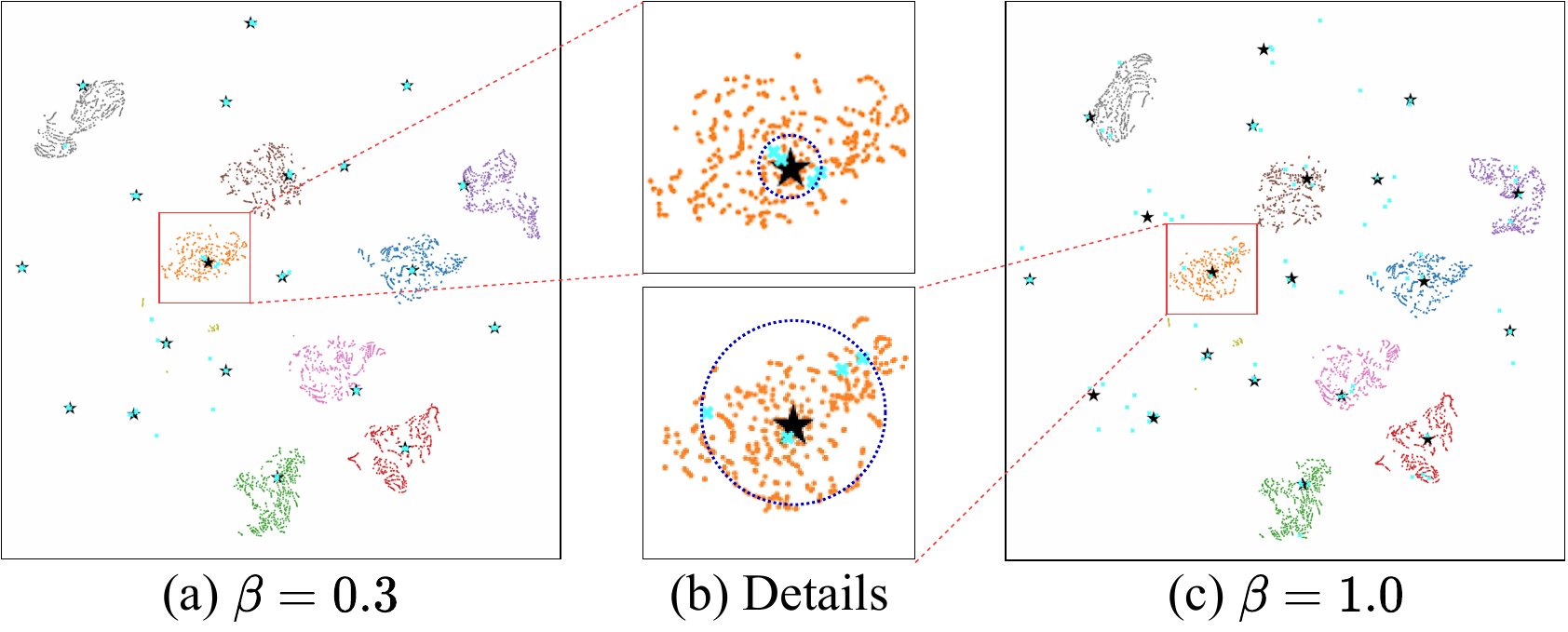}
  \caption{Visualization of virtual negatives.
   }
  \label{vis_vn}
\end{figure}

\noindent\textbf{Visualization of the virtual negatives.}
To investigate how virtual negatives work, we utilize t-SNE to visualize them.
Fig.~\ref{vis_vn} illustrates the results.
The VNs are observed to form clusters around the global distribution prototype, while the real representations also exhibit clustering around the global distribution prototype, but with notable displacements. 
This observation indicates that virtual negatives are more effective in capturing global features compared to real negatives, which primarily represent local-level features.
Furthermore, virtual negatives occupy positions that should have been filled by real negatives but were lost due to the limited capacity of the mini-batch
We visualize two sets of VNs with different virtual radius $\beta$, enabling a balance between the compactness and diversity of VNs, as Fig.~\ref{vis_vn} \textcolor{red}{(b)} details.

\begin{figure}[t]
  \centering
  \includegraphics[width=1.0\linewidth]{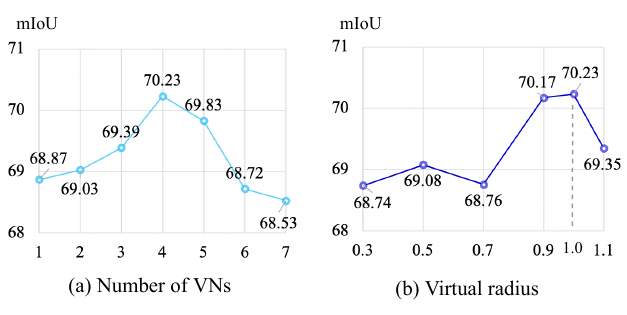}
  \caption{Effect on the number of virtual negatives and the virtual radius.
   }
  \label{vn}
\end{figure}
\noindent\textbf{Results of using virtual negatives.}
Our VN introduces two critical hyper-parameters: its number and virtual radius $\beta$.
Both of these parameters have a significant impact on the properties of the negative distribution and consequently affect the overall performance.
To investigate the influence of the number of VNs on performance, we conducted several experiments with varying numbers of VNs.
Fig.~\ref{vn} \textcolor{red}{(a)} demonstrates that using 4 VNs for each class performs best in the current setting.
We argue that an excessive number of VNs may excessively focus on the global representation while disregarding the local representation.
This imbalance may hinder local contrasts, thereby impeding the update of global distribution prototypes based on local prototypes.
On the other hand, too few VNs may not fully leverage the benefits of global representations, resulting in a diminished impact on the learning process.
To explore the impact of the virtual radius on performance, we conducted multiple experiments with different values of $\beta$.
As shown in Fig.~\ref{vn} \textcolor{red}{(b)}, our VN achieves optimal performance with a virtual radius of $\beta=1.0$.
The $\beta$ controls the diversity and noise in generated VNs, and there also will be a dilemma between diversity and noise in VNs.
Specifically, too large $\beta$ will introduce too much noise in VNs, even though it endows VNs with more diversity.
In contrast, too small $\beta$ loses diversity when reducing noise in VNs.
The number of VNs and the virtual radius both exert non-linear effects on the properties of the virtual negatives, which will influence the overall performance of the framework.
Overall, carefully selecting the number of VNs and the virtual radius is crucial to strike a balance between local and global representations, ultimately enhancing the performance of the model.

\noindent\textbf{Comparisons to the memory bank strategy.}
To compare different compensation strategies, we conduct experiments based on a single NVIDIA Tesla V100 GPU with three different strategies: no compensation strategy (w/o strategy), memory bank strategy (MB), and our virtual negatives (VN).
It is worth noting that we use our baseline framework with PR and GDP to conduct experiments.
To ensure a fair comparison and mitigate the impact of an inconsistent number of negative representations, we included an equal number of real representations in the experiments without the update strategy.
Since the memory bank strategy consumes a substantial amount of memory, necessitating the release of memory and reduction in batch size. Consequently, this leads to a degradation in the performance of pixel-wise contrastive learning.
As Tab.~\ref{tab6} shows, the experiment without any compensation strategy achieves $66.89\%$ mIoU.
Introducing the memory bank strategy improves the mIoU by $0.22\%$, while our VN strategy achieves a more substantial improvement of $\bm{1.12\%}$ mIoU.
It is important to consider the trade-off between the performance gain and the associated resource costs.
The memory bank strategy consumes a considerable amount of memory, approximately $2.63$ GB, while our VN strategy requires only $\bm{42}$ KB of memory.
Additionally, the training time for our VN strategy increases by only $\bm{0.03}$ GPU days, whereas the memory bank strategy requires an additional $0.47$ GPU days.
These results suggest that the memory bank strategy may not be the optimal solution for pixel-wise contrastive learning.
Due to the limited storage capacity, the number of representations stored in memory bank is limited and the update in memory bank is frequent.
Therefore, the memory bank approach discards most historical representations and loses global features.
In contrast, our VNs, generated by GDPs, consider all historical representations because the update strategy for GDP utilizes all representations in the training process.
As a result, our VNs are more effective in capturing global features.

\begin{table}[h]
\caption{Comparison under different compensation strategies. All experiments are performed on Pascal VOC 2012 with 92 labeled images. The measurement of times is GPU days.}
\setlength{\tabcolsep}{3.8mm}{%
\begin{tabular}{cc|c|c|c}
\hline
\multicolumn{2}{c|}{\multirow{2}{*}{w/o strategy}} & mIoU & Memory & Times \\ \cline{3-5} 
\multicolumn{2}{c|}{}                              & $66.89$     & $0$M       & $1.51$      \\ \hline
\multicolumn{1}{c|}{\multirow{3}{*}{MB}} & size    & mIoU & Memory & Times \\ \cline{2-5}
\multicolumn{1}{c|}{}                    & $30000$   & $66.05$     & $1.20$GB       & $1.67$      \\
\multicolumn{1}{c|}{}                    & $65536$   & $67.11$     & $2.63$GB       & $1.98$      \\ \hline
\multicolumn{1}{c|}{\multirow{6}{*}{VN}} & $\beta$ & mIoU & Memory & Times \\ \cline{2-5} 
\multicolumn{1}{c|}{}                    & $0.7$        & $66.95$     &        &       \\
\multicolumn{1}{c|}{}                    & $0.8$        & \bm{$68.01$}     & \multirow{4}{*}{$42$KB}       & \multirow{4}{*}{$1.54$}      \\
\multicolumn{1}{c|}{}                    & $0.9$        & $66.04$     &        &       \\
\multicolumn{1}{c|}{}                    & $1.0$        & $66.23$     &        &       \\
\multicolumn{1}{c|}{}                    & $1.1$        & $67.98$     &        &       \\
\multicolumn{1}{c|}{}                    & $1.2$        & $67.06$     &        &       \\ \hline
\end{tabular}
}
\label{tab6}
\end{table}

\subsection{Other Ablative Studies}
In this section, we conduct some other ablative studies to further explore our framework.

\begin{figure}[t]
  \centering
  \includegraphics[width=1.0\linewidth]{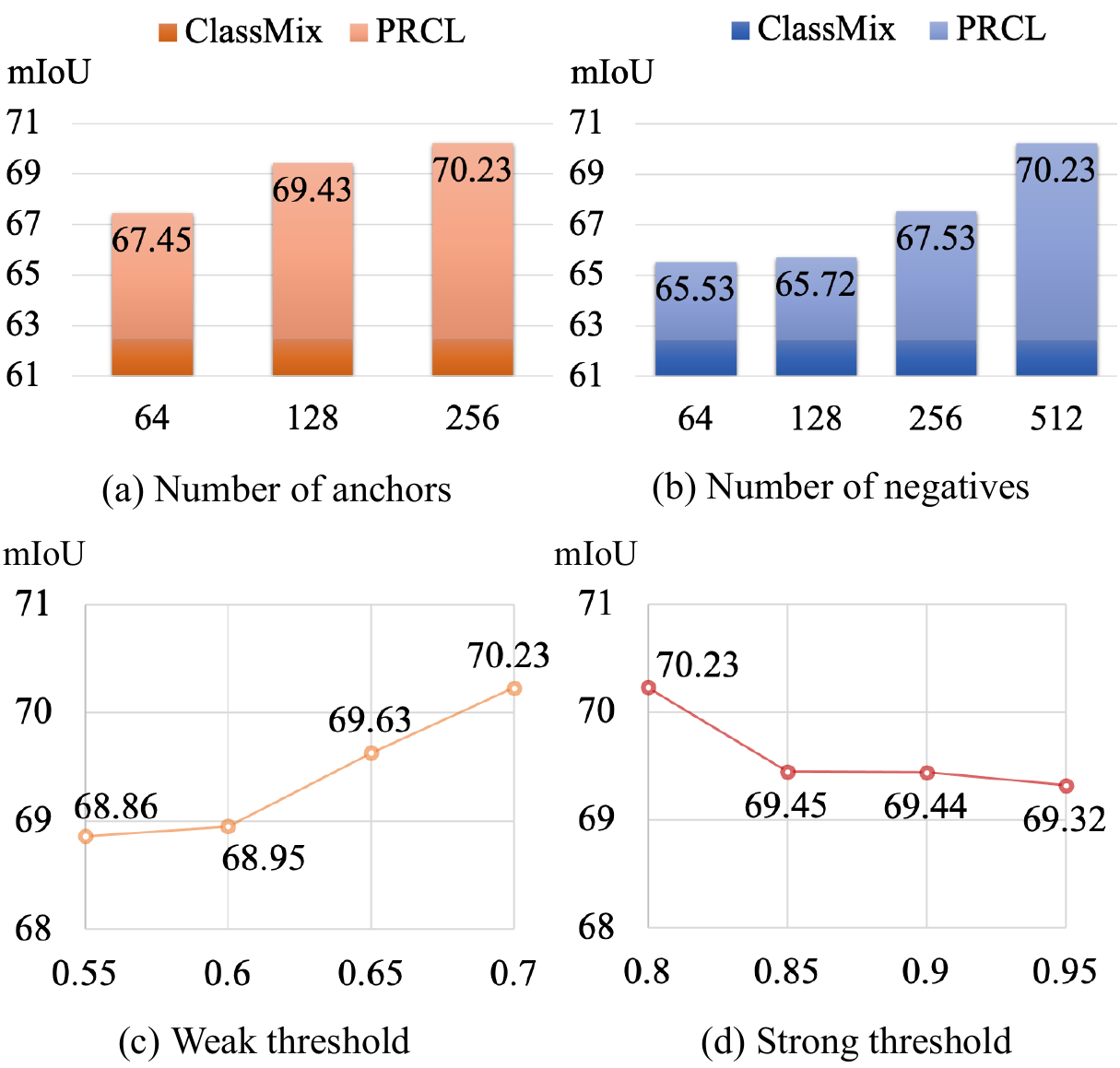}
  \caption{Ablation study on the number of samples and thresholds.
   }
  \label{sample_strategy}
\end{figure}

\noindent\textbf{Impact of sampling numbers.}
We conducted two sets of experiments to investigate the impact of different sampling numbers on the performance of our framework.
Firstly, we explored the effect of varying the number of anchors.
As Fig.~\ref{sample_strategy} \textcolor{red}{(a)} shows, increasing the number of anchors led to improved performance within the constraints of our limited GPU memory.
The number of anchors exhibited a significant influence on the overall performance, suggesting that a careful selection of anchor representations is crucial for achieving optimal results.
Similarly, we examined the influence of different numbers of negatives in our experiments. 
Fig.~\ref{sample_strategy} \textcolor{red}{(b)} illustrates that increasing the number of negatives resulted in improved performance.
This can be attributed to the fact that a larger set of negatives provides a more comprehensive representation of the ideal negative distribution, which in turn has a positive impact on the contrastive learning process.
These findings also support our motivation for compensating for the distribution of negative representations at a global level.

\noindent\textbf{Impact of thresholds.}
We conduct multiple experiments with different strong thresholds $\delta_s$ and weak thresholds $\delta_w$ by varying one threshold while keeping the other constant.
As shown in Fig.~\ref{sample_strategy} \textcolor{red}{(c)} and \textcolor{red}{(d)}, the strategy with $\delta_s=0.80$ and $\delta_w=0.70$ yields the best performance.
This outcome can be primarily attributed to the sampling of more ambiguous and challenging anchor representations achieved by using a lower $\delta_s$.
And we argue that more ambiguous and hard anchors are valuable in contrast since the model does not fully grasp the information of these corresponding pixels.

\begin{table}[t]
\centering
\caption{Ablation study on the effectiveness of components in our framework, including Probabilistic Representation (PR), Global Distribution Prototype (GDP) and Virtual Negatives (VN)}
\setlength{\tabcolsep}{5.0mm}{%
\begin{tabular}{c|cc}
\hline
 component  & mIoU & obtain \\ \hline
 baseline   & $66.91$  & -  \\
 PR         & $68.49$  & $1.58$  \\
 PR+GDP     & $69.52$  & $2.61$  \\
 PR+VN      & $69.07$  & $2.16$  \\
 PR+GDP+VN  & $70.23$  & $3.32$  \\ \hline
\end{tabular}
}
\label{tab10}
\end{table}

\noindent\textbf{Impact of components.}
We conduct experiments in Tab.~\ref{tab10} to ablate each component of our framework on PASCAL VOC 2012 with 92 labeled images.
Our baseline achieves mIoU of $66.91\%$.
Simply substituting vanilla representation for probabilistic representation (PR) improves the baseline by $1.58\%$ mIoU.
On the basis of this, we introduce the global distribution prototype (GDP), which additionally boosts the performance of $1.03\%$ mIoU.
This improvement demonstrates the necessity and effectiveness of prototype consistency.
Meanwhile, we introduce virtual negatives (VN), which compensate for the fragmentary negative distribution and include more global features in the contrast.
This strategy also boosts the performance of $0.58\%$ mIoU.
Finally, combining these two increases performance by $1.74\%$ mIoU.
Collectively, these experiments demonstrate the significant and distinct contributions of each component in our proposed method, culminating in improved performance and highlighting the effectiveness of our framework.

\section{Conclusion}
In this paper, we present a novel framework, termed PRCL, which enhances the robustness of contrastive learning by incorporating probabilistic representations, thus effectively addressing the challenges posed by inaccurate pseudo-labels.
Furthermore, our method introduces two key components, namely the global distribution prototype and the virtual negative, to overcome the limitations arising from the limited size of the mini-batch.
Comprehensive experiments conducted on various datasets validate the efficacy of our proposed components, as they significantly improve the model's robustness and enhance overall performance.


\clearpage
\appendix
\section{Proof of Equation 4}
Generally, we regard the prototype as the posterior distribution after the $n^{th}$ observations of representations $\{z_1, z_2, \dots, z_n\}$.
Meanwhile, we assume that all the observations are conditionally independent, the distribution prototype can be derived as $p(\bm{\rho}|\bm{z}_1, \bm{z}_2, ..., \bm{z}_{n})$.
Without loss of generality, we only consider a one-dimensional case here.
It is easy to extend the proof to all dimensions since each dimension of the feature is supposed to be independent.
We assume that the distribution prototype $p(\bm{\rho}|\bm{z}_1, \bm{z}_2, ..., \bm{z}_{n})$ is with $\hat{\mu}_n$ and $\hat{\sigma}^2_n$ as mean and variance, respectively.
Now we need to add a new representation as the observation to obtain a new prototype $p(\bm{\rho}|\bm{z}_1, \bm{z}_2, ..., \bm{z}_{n+1})$, if we take log on this prototype, we have:
\begin{equation}
    \scalebox{0.92}{$ \displaystyle
    \begin{aligned}
        &\log p(\bm{\rho}|\bm{z}_1, \bm{z}_2, ..., \bm{z}_{n+1}) \\
        =&\log p(\bm{\rho}|\bm{z}_{n+1})+\log p(\bm{\rho}|\bm{z}_1, \bm{z}_2, ..., \bm{z}_{n})-\log p(\bm{\rho}) + const \\
        =&-\frac{(\bm{\rho}-\bm{\mu}_{n+1})^2}{2\bm{\sigma}^2_{n+1}}-\frac{(\bm{\rho}-\hat{\bm{\mu}}_{n})^2}{2\hat{\bm{\sigma}}^2_{n}}+\frac{(\bm{\rho}-\bm{\mu}_{0})^2}{2\bm{\sigma}^2_{0}} + const \\
        =&-\frac{(\bm{\rho}-\hat{\bm{\mu}}_{n+1})^2}{2\hat{\bm{\sigma}}^2_{n+1}} + const,
    \end{aligned}
    $}
\end{equation}
where "const" means the constant which is irrelevant to the prototype $\bm{\rho}$
and
\begin{equation}
    \scalebox{0.92}{$ \displaystyle
    \begin{aligned}
        \hat{\bm{\mu}}_{n+1}=\hat{\bm{\sigma}}^2_{n+1}\Biggl( \frac{\bm{\mu}_{n+1}}{\bm{\sigma}^2_{n+1}} + \frac{\hat{\bm{\mu}}_{n}}{\hat{\bm{\sigma}}^2_{n}} - \frac{\bm{\mu}_{0}}{\bm{\sigma}^2_{0}}\Biggl),
    \end{aligned}
    $}
\end{equation}
\begin{equation}
    \scalebox{0.92}{$ \displaystyle
    \begin{aligned}
        \frac{1}{\hat{\bm{\sigma}}^2_{n+1}}=\frac{1}{\bm{\sigma}^2_{n+1}}+\frac{1}{\bm{\sigma}^2_{n}}-\frac{1}{\bm{\sigma}^2_{0}}.
    \end{aligned}
    $}
\end{equation}
$\bm{\sigma}_0$ is the $\bm{\sigma}$ of the first representation.
At the beginning of the training process, the representation is unreasonable, so we consider that the reliability is quite low and $\bm{\sigma}_0\rightarrow \infty$ (corresponds to an extremely large value in experiments).
We have
\begin{equation}
    \scalebox{0.92}{$ \displaystyle
    \begin{aligned}
        \hat{\bm{\mu}}_{n+1}=\frac{\hat{\bm{\sigma}}^2_{n}\bm{\mu}_{n+1}+\bm{\sigma}^2_{n+1}\hat{\bm{\mu}}_{n}}{\bm{\sigma}^2_{n+1}+\hat{\bm{\sigma}}^2_{n}},
    \end{aligned}
    $}
\end{equation}
\begin{equation}
    \scalebox{0.92}{$ \displaystyle
    \begin{aligned}
        \frac{1}{\hat{\bm{\sigma}}_{n+1}}=\frac{\bm{\sigma}^2_{n+1}+\hat{\bm{\sigma}}^2_{n}}{\bm{\sigma}^2_{n+1}\hat{\bm{\sigma}}^2_{n}}.
    \end{aligned}
    $}
\end{equation}
Above is the process of obtaining $\rho_{n+1}$ through $\rho_{n}$.
Next, we briefly give the solution of obtaining $\rho_{n+1}$ using $n+1$ representations.
\begin{equation}
    \scalebox{0.92}{$ \displaystyle
    \begin{aligned}
        & \log p\left(\rho \mid \bm{z}_1, \bm{z}_2, \ldots, \bm{z}_n\right) \\
        = & \log \left[\alpha p\left(\rho \mid \bm{z}_1\right) \prod_{i=2}^n \frac{p\left(\rho \mid \bm{z}_i\right)}{p(\rho)}\right] \\
        = & (n-1) \log p(\rho)-\sum_{i=1}^n \log p\left(\rho \mid \bm{z}_i\right)+ const \\
        = & (n-1) \frac{\left(\rho-\mu_0\right)^2}{2 \sigma_0^2}-\sum_{i=1}^n \frac{\left(\rho-\mu_i\right)^2}{2 \sigma_i^2}+ const \\
        = & -\frac{\left(\rho-\hat{\mu}_n\right)^2}{2 \hat{\sigma}_n^2} + const,
    \end{aligned}
    $}
\end{equation}
where $\alpha=\frac{\prod_{i=1}^n p(\bm{\rho}_i)}{p(\bm{z}_1, \bm{z}_2, ..., \bm{z}_{n})}$ and
\begin{equation}
    \scalebox{0.92}{$ \displaystyle
    \begin{aligned}
        \hat{\bm{\mu}}_n = \sum_{i=1}^n\frac{\hat{\bm{\sigma}}_n^2}{\bm{\sigma}_i^2}\bm{\mu}_i-(n-1)\frac{\hat{\bm{\sigma}}_n^2}{\bm{\sigma}_0^2}\bm{\mu}_0,
    \end{aligned}
    $}
\end{equation}
\begin{equation}
    \scalebox{0.92}{$ \displaystyle
    \begin{aligned}
        \frac{1}{\hat{\bm{\sigma}}^2} = \sum_{i=1}^n\frac{1}{\bm{\sigma}_i^2}-(n-1)\frac{1}{\bm{\sigma}_0^2}.
    \end{aligned}
    $}
\end{equation}
Because $\bm{\sigma}_0\rightarrow \infty$, we have
\begin{equation}
    \scalebox{0.92}{$ \displaystyle
    \begin{aligned}
        \hat{\bm{\mu}}_n = \sum_{i=1}^n\frac{\hat{\bm{\sigma}}_n^2}{\bm{\sigma}_i^2}\bm{\mu}_i,
    \end{aligned}
    $}
\end{equation}
\begin{equation}
    \scalebox{0.92}{$ \displaystyle
    \begin{aligned}
        \frac{1}{\hat{\bm{\sigma}}^2} = \sum_{i=1}^n\frac{1}{\bm{\sigma}_i^2}.
    \end{aligned}
    $}
\end{equation}

\bibliographystyle{spmpsci} 
\bibliography{sn-bibliography}

\end{document}